%% file: neurips_2025.tex
\documentclass{article}

\usepackage[final]{neurips_2025}

\usepackage[utf8]{inputenc} 
\usepackage[T1]{fontenc}    
\usepackage{hyperref}       
\usepackage{url}            
\usepackage{booktabs}       
\usepackage{amsfonts}       
\usepackage{nicefrac}       
\usepackage{microtype}      
\usepackage{xcolor}         

\usepackage{algorithm}
\usepackage{algpseudocode}
\usepackage{makecell}
\usepackage{pgfplots}
\usepackage[svgnames]{xcolor}
\usepackage{tablefootnote}
\pgfplotsset{compat=1.18}
\usetikzlibrary{
    arrows,
    arrows.meta,
    backgrounds,
    bending,
    calc,
    fit,
    patterns,
    positioning,
    shapes,
    shapes.callouts,
    shapes.multipart
}

\title{Loquetier: A Virtualized Multi-LoRA Framework\\ for Unified LLM Fine-tuning and Serving}

\author{
Yuchen Zhang$^1$ \quad Hanyue Du$^1$ \quad Chun Cao$^1$ \quad Jingwei Xu$^{1}$\thanks{Corresponding author} \\
$^1$State Key Laboratory for Novel Software Technology, Nanjing University, China\\
\texttt{\{zfirozen,dhy\}@smail.nju.edu.cn, \{caochun,jingweix\}@nju.edu.cn}
}

\begin{document}

\maketitle

\begin{abstract}

Low-Rank Adaptation (LoRA) has become a widely adopted parameter-efficient fine-tuning (PEFT) technique for adapting large language models (LLMs) to downstream tasks. While prior work has explored strategies for integrating LLM training and serving, there still remains a gap in unifying fine-tuning and inference for LoRA-based models. We present \textbf{Loquetier}, a virtualized multi-LoRA framework that seamlessly integrates LoRA fine-tuning and serving within a single runtime. Loquetier introduces two key components: (1) a \emph{Virtualized Module} that isolates PEFT-based modifications and supports multiple adapters on a shared base model, and (2) an optimized computation flow with a kernel design that merges fine-tuning and inference paths in forward propagation, enabling efficient batching and minimizing kernel invocation overhead. Extensive experiments across three task settings show that Loquetier consistently outperforms existing baselines in both performance and flexibility, achieving up to $3.0\times$ the throughput of the state-of-the-art co-serving system on inference-only tasks and $46.4\times$ higher SLO attainment than PEFT on unified fine-tuning and inference tasks. The implementation of Loquetier is publicly available at \url{https://github.com/NJUDeepEngine/Loquetier}.
\end{abstract}

\input{chapters/chapter1}

\input{chapters/chapter2}

\input{chapters/chapter3}

\input{chapters/chapter4}

\input{chapters/chapter5}

\begin{ack}
We are thankful to the anonymous reviewers for their helpful comments.
This work is supported by Frontier Technologies R\&D Program of Jiangsu (\#BF2024059), the National Natural Science Foundation of China (Grants \#62172199), the Collaborative Innovation
Center of Novel Software Technology and Industrialization, and Jiangsu Wukong Intelligent Computing Digital Technology. Jingwei Xu (\texttt{jingweix@nju.edu.cn}) is the corresponding author. 
\end{ack}


\newpage
\bibliographystyle{unsrtnat}
\bibliography{references}


\input{chapters/appendix}


\newpage
\section*{NeurIPS Paper Checklist}

The checklist is designed to encourage best practices for responsible machine learning research, addressing issues of reproducibility, transparency, research ethics, and societal impact. Do not remove the checklist: {\bf The papers not including the checklist will be desk rejected.} The checklist should follow the references and follow the (optional) supplemental material.  The checklist does NOT count towards the page
limit. 

Please read the checklist guidelines carefully for information on how to answer these questions. For each question in the checklist:
\begin{itemize}
    \item You should answer \answerYes{}, \answerNo{}, or \answerNA{}.
    \item \answerNA{} means either that the question is Not Applicable for that particular paper or the relevant information is Not Available.
    \item Please provide a short (1–2 sentence) justification right after your answer (even for NA). 
\end{itemize}

{\bf The checklist answers are an integral part of your paper submission.} They are visible to the reviewers, area chairs, senior area chairs, and ethics reviewers. You will be asked to also include it (after eventual revisions) with the final version of your paper, and its final version will be published with the paper.

The reviewers of your paper will be asked to use the checklist as one of the factors in their evaluation. While "\answerYes{}" is generally preferable to "\answerNo{}", it is perfectly acceptable to answer "\answerNo{}" provided a proper justification is given (e.g., "error bars are not reported because it would be too computationally expensive" or "we were unable to find the license for the dataset we used"). In general, answering "\answerNo{}" or "\answerNA{}" is not grounds for rejection. While the questions are phrased in a binary way, we acknowledge that the true answer is often more nuanced, so please just use your best judgment and write a justification to elaborate. All supporting evidence can appear either in the main paper or the supplemental material, provided in appendix. If you answer \answerYes{} to a question, in the justification please point to the section(s) where related material for the question can be found.

IMPORTANT, please:
\begin{itemize}
    \item {\bf Delete this instruction block, but keep the section heading ``NeurIPS Paper Checklist"},
    \item  {\bf Keep the checklist subsection headings, questions/answers and guidelines below.}
    \item {\bf Do not modify the questions and only use the provided macros for your answers}.
\end{itemize}


\begin{enumerate}

\item {\bf Claims}
    \item[] Question: Do the main claims made in the abstract and introduction accurately reflect the paper's contributions and scope?
    \item[] Answer: \answerYes{} 
    \item[] Justification: The paper proposes a redesigned kernel SMLM and a unified computation flow for the unified operation of fine-tuning and inference tasks for LoRA models (Section~\ref{sec:compflow-smlm-kernel}), and Virtualized Module implementation to isolate PEFT-based model modifications and flexible instance-to-instance migration (Section~\ref{sec:virtual-module}).

\item {\bf Limitations}
    \item[] Question: Does the paper discuss the limitations of the work performed by the authors?
    \item[] Answer: \answerYes{} 
    \item[] Justification: The limitations of the work is discussed in Section \ref{sec:limitations}.

\item {\bf Theory assumptions and proofs}
    \item[] Question: For each theoretical result, does the paper provide the full set of assumptions and a complete (and correct) proof?
    \item[] Answer: \answerNA{} 
    \item[] Justification: This paper does not include theoretical results.

    \item {\bf Experimental result reproducibility}
    \item[] Question: Does the paper fully disclose all the information needed to reproduce the main experimental results of the paper to the extent that it affects the main claims and/or conclusions of the paper (regardless of whether the code and data are provided or not)?
    \item[] Answer: \answerYes{} 
    \item[] Justification: The paper discloses all the information needed for reproducibility (Section \ref{sec:exp-settings}). This paper releases the code for quick reproducibility. The implementation code is available at \url{https://github.com/s3co3wjy5tr2bdfj/Loquetier}.

\item {\bf Open access to data and code}
    \item[] Question: Does the paper provide open access to the data and code, with sufficient instructions to faithfully reproduce the main experimental results, as described in supplemental material?
    \item[] Answer: \answerYes{} 
    \item[] Justification: The implementation code is available at \url{https://github.com/s3co3wjy5tr2bdfj/Loquetier}.

\item {\bf Experimental setting/details}
    \item[] Question: Does the paper specify all the training and test details (e.g., data splits, hyperparameters, how they were chosen, type of optimizer, etc.) necessary to understand the results?
    \item[] Answer: \answerYes{} 
    \item[] Justification: The experimental settings are presented in Section \ref{sec:exp-settings}. More details could be found in our code repository, which is available at \url{https://github.com/s3co3wjy5tr2bdfj/Loquetier}.

\item {\bf Experiment statistical significance}
    \item[] Question: Does the paper report error bars suitably and correctly defined or other appropriate information about the statistical significance of the experiments?
    \item[] Answer: \answerNo{} 
    \item[] Justification: The experimental metrics are already the statistical result of a large amount of data, so there is no need to conduct multiple experiments to plot the errorbar. For the data in Figure~\ref{fig:test-unified}, we took a total of 5 data before and after each data point for smoothing operations to improve chart readability and eliminate extreme values over short periods of time.

\item {\bf Experiments compute resources}
    \item[] Question: For each experiment, does the paper provide sufficient information on the computer resources (type of compute workers, memory, time of execution) needed to reproduce the experiments?
    \item[] Answer: \answerYes{} 
    \item[] Justification: We present the information on the computer resources in Section \ref{sec:exp-settings}. 
    That is, we test the inference-only tasks on a server with 4 NVIDIA A6000 48G GPUs. We test the fine-tuning-only tasks and the unified fine-tuning and inference tasks on servers with 4 NVIDIA H800 80G GPUs. Each test process has at least 128G of host memory available.
    
\item {\bf Code of ethics}
    \item[] Question: Does the research conducted in the paper conform, in every respect, with the NeurIPS Code of Ethics \url{https://neurips.cc/public/EthicsGuidelines}?
    \item[] Answer: \answerYes{} 
    \item[] Justification: We conformed the NeurIPS Code of Ethics and make sure to preserve anonymity.

\item {\bf Broader impacts}
    \item[] Question: Does the paper discuss both potential positive societal impacts and negative societal impacts of the work performed?
    \item[] Answer: \answerNA{} 
    \item[] Justification: There is no societal impact of the work performed. Our work proposes a framework for unified LLM fine-tuning and serving on multiple LoRA models, which does not lead to any negative societal impacts.
    
\item {\bf Safeguards}
    \item[] Question: Does the paper describe safeguards that have been put in place for responsible release of data or models that have a high risk for misuse (e.g., pretrained language models, image generators, or scraped datasets)?
    \item[] Answer: \answerNA{} 
    \item[] Justification: This paper presents a framework for unified LLM fine-tuning and serving with an SMLM kernel for unified tasks, an unified computation flow management, and an implementation of Virtualized Module. This framework does not include any data or models, and therefore this work does not pose any associated risks.

\item {\bf Licenses for existing assets}
    \item[] Question: Are the creators or original owners of assets (e.g., code, data, models), used in the paper, properly credited and are the license and terms of use explicitly mentioned and properly respected?
    \item[] Answer: \answerYes{} 
    \item[] Justification: All existing assets used in the paper are listed here. Alpaca: Creative Commons Attribution Non Commercial 4.0, \url{https://huggingface.co/datasets/tatsu-lab/alpaca}. GSM8K: MIT License, \url{https://huggingface.co/datasets/openai/gsm8k}. Llama3-8B: Llama 3 Community License Agreement, \url{https://huggingface.co/meta-llama/Meta-Llama-3-8B}. ShareGPT Vicuna: Apache License Version 2.0, \url{https://huggingface.co/datasets/anon8231489123/ShareGPT_Vicuna_unfiltered}.

\item {\bf New assets}
    \item[] Question: Are new assets introduced in the paper well documented and is the documentation provided alongside the assets?
    \item[] Answer: \answerYes{} 
    \item[] Justification: Our released code is well organized in the repository.

\item {\bf Crowdsourcing and research with human subjects}
    \item[] Question: For crowdsourcing experiments and research with human subjects, does the paper include the full text of instructions given to participants and screenshots, if applicable, as well as details about compensation (if any)? 
    \item[] Answer: \answerNA{} 
    \item[] Justification: This paper does not involve crowdsourcing nor research with human subjects.

\item {\bf Institutional review board (IRB) approvals or equivalent for research with human subjects}
    \item[] Question: Does the paper describe potential risks incurred by study participants, whether such risks were disclosed to the subjects, and whether Institutional Review Board (IRB) approvals (or an equivalent approval/review based on the requirements of your country or institution) were obtained?
    \item[] Answer: \answerNA{} 
    \item[] Justification: This paper does not involve crowdsourcing nor research with human subjects.

\item {\bf Declaration of LLM usage}
    \item[] Question: Does the paper describe the usage of LLMs if it is an important, original, or non-standard component of the core methods in this research? Note that if the LLM is used only for writing, editing, or formatting purposes and does not impact the core methodology, scientific rigorousness, or originality of the research, declaration is not required.
    \item[] Answer: \answerNA{} 
    \item[] Justification: The entire approach development in this research does not involve LLMs as any important, original, or non-standard components.

\end{enumerate}

\end{document}

%% file: chapters/chapter1.tex
\section{Introduction}
Large Language Models (LLMs) built on stacked transformer blocks \citep{vaswani_attention_2017} have achieved remarkable success across a wide range of text generation tasks. This success has driven the development of ever-larger models, such as the LlaMA series \citep{touvron_llama_2023, grattafiori_llama_2024} and the Qwen family \citep{bai_qwen_2023, yang_qwen2_2024}. However, the rapid growth in model size has introduced prohibitive costs. For example, LlaMA~3 contains 405B parameters, and DeepSeek-V3 \citep{bi_deepseek_2024} scales to 671B. Their computational and memory requirements of full-parameter training now represent a major bottleneck, restricting both the scalability and accessibility of LLM development.

Parameter-efficient fine-tuning (PEFT) has emerged as a practical solution to these challenges. By reducing the number of trainable parameters while retaining the effectiveness of full-model fine-tuning, PEFT offers a balance between efficiency and adaptability \citep{ding_parameter-efficient_2023}. Recent studies have evaluated PEFT across diverse applications and theoretical dimensions \citep{balne_parameter_2024, xu_parameter-efficient_2023}, and have surveyed its underlying mechanisms and practical benefits \citep{han_parameter-efficient_2024, fu_effectiveness_2023}. Notice that PEFT approaches such as Prefix Tuning \citep{li_prefix-tuning_2021} and Prompt Tuning \citep{lester_power_2021} have demonstrated strong adaptability by optimizing small task-specific vectors. Moreover, PEFT has shown competitive or superior performance in low-resource settings, including zero- and few-shot learning \citep{liu_few-shot_2022, hu_llm-adapters_2023}.

Among PEFT approaches, \emph{Low-Rank Adaptation} (LoRA) \citep{hu_lora_2022} has become particularly prominent, offering scalable and effective fine-tuning across diverse LLM tasks. Numerous extensions have been proposed to enhance its flexibility, including LoHa \citep{hyeon-woo_fedpara_2021}, VeRA \citep{kopiczko_vera_2023}, and LoKr \citep{yeh_navigating_2023}. LoRA has also shown strong potential for personalization, powering systems in recommendation \citep{kong_customizing_2024, zhu_lifelong_2024}, and user-centered content generation \citep{zhang_personalized_2024, wu_difflora_2024}. Its modularity makes it well-suited for large-scale serving systems where real-time customization is critical.

In practice, systems that serve LoRA often need to support fine-tuning adapters while simultaneously deploying them for inference across diverse tasks. However, no existing framework can seamlessly unify these two capabilities, resulting major obstacles for scaling LoRA into production. Jointly fine-tuning and serving multiple adapters requires minimizing memory and computation overhead while efficiently handling heterogeneous workloads. Prior efforts have mainly optimized the base LLM, such as through KVCache improvements or inference pipeline parallelization \citep{li_llm_2024}, with only limited advances in multi-LoRA inference \citep{chen_punica_2024, sheng_s-lora_2023} or integrated fine-tuning and inference \citep{miao_flexllm_2024}. However, these approaches still face critical challenges: adapters are often fused into monolithic instances for efficiency, thus they cannot be dynamically loaded or unloaded; decoding efficiency degrades significantly when fine-tuning and inference run concurrently; and task switching typically requires halting the current job before starting another, causing downtime, bandwidth overhead, and wasted resources. As a result, existing frameworks remain inadequate for large-scale, production-ready LoRA applications.

In this paper, we introduce \textbf{Loquetier}\footnote{The name \emph{Loquetier} is a synthesis of ``LoRA'' and ``coquetier'', reflecting our design philosophy: the base model serves as the foundational spirit, while LoRA modules act as adjunct ingredients-spirits, juices, and syrups-mixed in for task-specific customization.}, a unified virtualization framework that integrates fine-tuning and serving of LLMs with LoRA-based PEFT. Loquetier provides a streamlined computation flow that handles both fine-tuning and inference requests within a shared runtime, using kernel-level optimizations to reduce memory overhead and execution latency. Specifically, Loquetier introduces the \emph{Segmented Multi-LoRA Multiplication (SMLM)} kernel, which enables mixed-task execution by distinguishing forward-pass behaviors for fine-tuning, evaluation, prefilling, and decoding. To support multiple concurrent LoRA adapters without modifying the base model, Loquetier further incorporates a \emph{Virtualized Module} abstraction that dynamically injects adapter logic while preserving compatibility and isolation across tasks and devices. This design enables seamless co-serving of heterogeneous LoRA configurations and supports instance-to-instance migration of fine-tuning jobs without kernel restarts or memory duplication. The main contributions are as follows:
\begin{itemize}
\item We design an SMLM kernel and an unified computation flow that efficiently supports fine-tuning and inference with multiple LoRA adapters on a shared base model.
\item We propose a modular virtualization mechanism that isolates PEFT-based modifications form the base model, enabling flexible instance-level migration and seamless adapter management.
\item We develop Loquetier to unify LoRA fine-tuning and serving. Extensive experiments show that Loquetier outperforms existing systems across diverse scenarios and enables unified practical fine-tuning and serving configurations previously unsupported.

\end{itemize}

\begin{table}
    \caption{Comparison on different LoRA tasks between Loquetier, PEFT and FlexLLM}
    \label{tab:tasks-comp}
    \centering
    \begin{tabular}{lllllll}
        \toprule
        & \multicolumn{2}{c}{Inference} & \multicolumn{2}{c}{Finetune} &
        \multicolumn{2}{c}{Finetune \& Inference}                 \\
        \cmidrule(r){2-3} \cmidrule(r){4-5} \cmidrule(r){6-7}
        Framework or System &
        Single & Multi & Single & Multi & Single & Multi          \\
        \midrule
        Loquetier &
        $\checkmark$ & $\checkmark$ & $\checkmark$ &
        $\checkmark$ & $\checkmark$ & $\checkmark$                \\
        PEFT &
        $\checkmark$ & $\checkmark$ & $\checkmark$ &
        $\times$     & $\checkmark$ & $\times$                    \\
        S-LoRA+PEFT &
        $\checkmark$ & $\checkmark$ & $\checkmark$ &
        $\times$     & $\checkmark$ & $\times$                    \\
        FlexLLM &
        $\checkmark$ &
        $\triangle$\tablefootnote{FlexLLM cycles through loading LoRA models during multi-LoRA inference, disregarding the maximum number of resident LoRAs set, which makes its multi-LoRA inference efficiency practically unusable.} &
        $\checkmark$\tablefootnote{The backward procedure of FlexLLM triggered an error originating from an unsupported operation in its gradient computation logic. We describe our solution in the Appendix~\ref{sec:solutions}.\label{fnote:flex-issue}} &
        $\times$\textsuperscript{\ref{fnote:flex-issue}} & $\times$ & $\times$            \\
        \bottomrule
    \end{tabular}
\end{table}

%% file: chapters/chapter2.tex
\section{Related Work}

\textbf{LLM inference optimization via KV cache.}
Key-value (KV) caching is a core technique for accelerating LLM inference by avoiding redundant computation and memory transfers \citep{pope_efficiently_2023}. Recent work improves cache efficiency and reduces GPU memory overhead through tailored management strategies. Prompt Cache \citep{gim_prompt_2024} reuses prompt embeddings to reduce duplication. vLLM \citep{kwon_efficient_2023} and vAttention \citep{prabhu_vattention_2025} address fragmentation using paging and virtual memory, respectively. Infinite-LLM \citep{lin_infinite-llm_2024} enables dynamic sharing of cache segments between host and GPU. LoongServe \citep{wu_loongserve_2024} enhances long-context serving by balancing prefilling and decoding workloads.

\textbf{LLM inference parallelism.}
A large amount of work improves LLM inference throughput by optimizing batch scheduling and pipeline execution. Response Length Perception \citep{zheng_response_2023} and \textbf{$S^3$} \citep{jin_s_2023} predict output lengths to batch similar requests, with $S^3$ further refining its predictions over time. Orca \citep{yu_orca_2022} uses token-level continuous batching, while DeepSpeed-Fastgen \citep{holmes_deepspeed-fastgen_2024} adjusts request lengths for better GPU utilization. Sarathi-Serve \citep{agrawal_taming_2024} co-schedules prefilling and decoding tasks, whereas TetriInfer \citep{hu_inference_2024}, Splitwise \citep{patel_splitwise_2024}, and DistServe \citep{zhong_distserve_2024} decouple these phases across threads, machines, or clusters to improve parallel efficiency.
In kernel, FlashDecoding++ \citep{hong_flashdecoding_2024} overlaps computation with data transfer to hide latency. FlashAttention \citep{dao_flashattention-2_2023, shah_flashattention-3_2024} uses warp specialization and asynchronous execution to maximize attention-layer throughput.

\textbf{Efficient multi-LoRA inference and unified fine-tuning-inference systems.}
FlashInfer \citep{ye_flashinfer_2025} reduces redundant storage and optimizes GPU memory access for efficient multi-LoRA inference. Cutlass \citep{thakkar_cutlass_2023}, a high-performance CUDA library for GEMM operations, serves as the kernel foundation for Punica \citep{chen_punica_2024}, which combines with FlashInfer to support scalable LoRA serving. Built on this stack, S-LoRA \citep{sheng_s-lora_2023} further improves GPU memory utilization through dynamic host-device memory transfers. In parallel, FlexLLM \citep{miao_flexllm_2024} explores unified token-level computation for co-serving inference and PEFT fine-tuning, though scalability remains limited.

\textbf{LoRA optimization and variants.}
There are lots of extensions that have been proposed to enhance the flexibility, efficiency, and adaptability of LoRA. AdaLoRA \citep{zhang_adalora_2023} introduces singular value decomposition for dynamic pruning, while IncreLoRA \citep{zhang_increlora_2023} allocates parameters based on module importance. SoRA \citep{ding_sparse_2023} adaptively adjusts rank via gated weight control during training. DiffoRA \citep{jiang_diffora_2025} selects modules to finetune using a Differential Adaptation Matrix (DAM). 
To improve expressiveness and stability, DoRA \citep{liu_dora_2024} decomposes weights into magnitude and direction. VB-LoRA \citep{li_vb-lora_2024} reduces redundancy by sharing global vector banks across modules. LoRA-XS \citep{balazy_lora-xs_2024} compresses storage with minimal $r \times r$ matrices, and QA-LoRA \citep{xu_qa-lora_2023} integrates quantization and grouping to increase representational flexibility.

%% file: chapters/chapter3.tex
\section{Loquetier Framework}
\input{figures/figure1}
In this section, we describe the overall architecture of Loquetier, followed by details of its unified computation flow and the proposed LoRA model virtualization.

\subsection{Framework}
As illustrated in Figure~\ref{fig:framework-struct}, Loquetier framework consists of two core components: (1) a model library centered around the \emph{Virtualized Module} for isolating PEFT modifications, and (2) a redesigned kernel and computation flow that jointly support fine-tuning and inference workloads. To enable concurrent execution, Loquetier integrates a context management system that coordinates the runtime scheduling of heterogeneous tasks.

When loading a base model into CPU or GPU memory, Loquetier instantiates multiple \emph{virtual models}, each acting as an isolated container for a specific PEFT configuration. These virtual models are bound to distinct adapters, enabling independent and concurrent execution. For LoRA-based adapters, we introduce the \texttt{MixedLoraModel} class to support fine-tuning. Based on our computation flow (see Section~\ref{sec:compflow-smlm-kernel}), each \texttt{MixedLoraModel} efficiently fine-tunes its associated LoRA adapter within its own container from dedicated trainers, while other virtual models can simultaneously remain in inference mode.

\begin{algorithm}[!tb]
    \renewcommand{\algorithmicrequire}{\textbf{Input:}}
    \renewcommand{\algorithmicensure}{\textbf{Output:}}
    \caption{Computation flow control in attention layer of Loquetier}
    \label{alg:comp-flow-attn}
    \begin{algorithmic}
        \Require
            hidden states matrix $\mathbf{X}$ with shape $[S, H]$,
            list of fine-tuning inputs batch-sequence information tuples $\mathbf{F}$,
            list of prefilling and evaluation inputs sequence lengths $\mathbf{P}$,
            decoding inputs count $\mathbf{D}$.
        \Ensure attention outputs matrix $\mathbf{O}$.
        \State $\mathbf{Q} = Q_{proj}(\mathbf{X})$; $\mathbf{K} = K_{proj}(\mathbf{X})$; $\mathbf{V} = V_{proj}(\mathbf{X})$;
        \State $\mathbf{Os} = [\space]$;
        \If{$len(\mathbf{F}) > 0$}
            \State Extract $\mathbf{Q_f}, \mathbf{K_f}, \mathbf{V_f}$ from $\mathbf{Q}, \mathbf{K}, \mathbf{V}$ based on $\mathbf{F}$;
            \State Compute $\mathbf{O_f}$ through the standard forward implementation;
            \State $\mathbf{O_f}$ is appended to $\mathbf{Os}$;
        \EndIf
        \If{$len(\mathbf{P}) > 0$}
            \State Compute offset of prefills $\mathbf{Offset_p}$ based on $\mathbf{F}$;
            \State Extract $\mathbf{Q_p}, \mathbf{K_p}, \mathbf{V_p}$ from $\mathbf{Q}, \mathbf{K}, \mathbf{V}$ based on $\mathbf{P}$ and $\mathbf{Offset_p}$;
            \State Initialize KVCache for prefills;
            \State Compute $\mathbf{O_p}$ through the FlashInfer forward implementation;
            \State $\mathbf{O_p}$ is appended to $\mathbf{Os}$;
        \EndIf
        \If{$\mathbf{D} > 0$}
            \State Compute offset of decodes $\mathbf{Offset_d}$ based on $\mathbf{F}$ and $\mathbf{P}$;
            \State Extract $\mathbf{Q_d}, \mathbf{K_d}, \mathbf{V_d}$ from $\mathbf{Q}, \mathbf{K}, \mathbf{V}$ based on $\mathbf{Offset_d}$;
            \State Append KVCache for decodes;
            \State Compute $\mathbf{O_d}$ through the standard forward implementation;
            \State $\mathbf{O_d}$ is appended to $\mathbf{Os}$;
        \EndIf
        \State Concatenate $\mathbf{Os}$ into one tensor as $\mathbf{O}$;
        \State $\mathbf{O} = O_{proj}(\mathbf{O})$;
        \State return $\mathbf{O}$;
    \end{algorithmic}
\end{algorithm}

Loquetier’s modular design is compatible with most existing LLM inference optimizations and training strategies. It fully supports architectures that leverage FlashInfer \citep{ye_flashinfer_2025} kernel and remains extensible to those that do not, requiring only localized modification to the inference logic within the computation flow. Section~\ref{sec:virtual-module} further details how the Virtualized Module enables base model sharing across diverse PEFT methods beyond LoRA.

\subsection{Virtualized module} 
\label{sec:virtual-module}

The mixing of different LoRA or other PEFT methods in the same model object makes model configurations chaotic and difficult to handle. Moreover, dynamic model loading and unloading should be supported in order to apply the fine-tuned and up-to-date LoRA models quickly.

We propose the Virtualized Module that provides methods and data proxies to foundation modules to solve the above problems. Applying the Virtualized Module to the base model is extremely low-cost, with no additional GPU memory overhead and provides independent model instances in an intuitive way. For each module type, the Virtualized Module creates the corresponding virtual module at runtime to provide the correct proxy methods. The Virtualized Module prepares additional properties and methods for Linear and Model to accommodate the architecture of Transformers and PEFT.

Since each virtual module class is created at runtime, which means none of these virtual modules can be shared, nor can a process that owns any virtual module create child processes from fork or spawn method. Furthermore, any deep copying behavior that includes a virtual module will cause the base module linked to it to be copied, thus invalidating base module sharing. For this reason, we provide a non-local class definition for deep copying, serialization, and deserialization. By voiding the containing Virtualized Module, LoRA models and other PEFT models loaded onto the Virtualized Module can be migrated to other GPU devices after deep-copying and used after unvoiding based on instances of the new Virtual Model.

Based on the above design, the Virtualized Module is compatible with all PEFT methods, as well as any custom model modification methods that do not modify the underlying model's own data, such as weights and module configurations. For scenarios where the target module overwrites the base module with new data, the module design needs to be normalized so that the target module runs the new forward method based on its own data and the base module's data, rather than running the base module's forward method after using destructive modification methods.

\subsection{Unified computation flow management and SMLM kernel}
\label{sec:compflow-smlm-kernel}

For multiple LoRA adapters within the same linear layer, traditional methods typically process the computation sequentially, computing the output for one LoRA at a time and iterating through all adapters. This approach significantly slows computation. Since different LoRAs within the same layer usually share identical or similar shapes, it is feasible to compute all outputs in a single kernel call. Punica \citep{chen_punica_2024} leverages this property by implementing a foundational algorithm on top of the Cutlass GEMM (General Matrix Multiplication) library \citep{thakkar_cutlass_2023}, enabling simultaneous processing of multiple input-LoRA pairs for efficient multi-adapter computation.

However, the original Punica kernel design is incompatible with fine-tuning, as it statically concatenates LoRA weights from the same module across different layers. This rigid coupling limits architectural flexibility and prevents selective application of LoRA to specific layers, which is particularly problematic in training scenarios where layerwise heterogeneity is common. This requires every layer in fine-tuning tasks to adopt the exact same configuration, even when certain linear modules do not actually require LoRA. Meanwhile, this design incurs substantial overhead in tracking and memory management, especially when operating on very large tensors during training. For inference tasks, it also eliminates the possibility of rapidly swapping LoRAs during runtime. Thus, the computation process must be halted before replacement, and the required LoRA must be re-spliced together.

To overcome these limitations, we adapt the Punica kernel to process LoRA weights one linear layer at a time. This decoupling removes the need to regenerate model files via weight transformation before execution. For static scaling factors in LoRA, we apply the scale directly to the weight tensor at \texttt{MixedLoraModel} instantiation to reduce runtime overhead. When dynamic scaling is required, it is applied on a per-request basis during the forward pass. Loquetier then executes forward computation across all active requests in a unified manner, supporting four types of requests: fine-tuning (training), evaluation, prefilling, and decoding. Evaluation requests are structurally similar to prefilling but compute a loss over labels and execute only a single generation pass, while prefilling requests omit the loss computation and transition into decoding after the initial pass.

\begin{algorithm}[!tb]
    \renewcommand{\algorithmicrequire}{\textbf{Input:}}
    \renewcommand{\algorithmicensure}{\textbf{Output:}}
    \caption{Computation flow control in causal LM of Loquetier}
    \label{alg:comp-flow-model}
    \begin{algorithmic}
        \Require
            LM inputs $\mathbf{X}$,
            list of fine-tuning inputs batch-sequence information tuples $\mathbf{F}$,
            list of evaluation inputs sequence lengths $\mathbf{E}$,
            Labels of fine-tuning and evaluation inputs $\mathbf{Labels}$,
            Accumulation steps of fine-tuning and evaluation inputs $\mathbf{A}$.
        \Ensure
            list of losses $\mathbf{Loss}$,
            logits $\mathbf{Logits}$.
        \State Compute $\mathbf{Logits}$ by forward propagation;
        \State $\mathbf{Loss} = [\space]$
        \For{(batch-sequence $\mathbf{(B, S)}$, accumulation step $\mathbf{A_{FE}}$ in $(\mathbf{F}, \mathbf{E}), \mathbf{A}$)}
            \State Extract $\mathbf{Logits_{FE}}$ and $\mathbf{Labels_{FE}}$ from $\mathbf{Logits}$ and $\mathbf{Labels}$;
            \State Shift $\mathbf{Logits_{FE}}$ and $\mathbf{Labels_{FE}}$;
            \State Compute $\mathbf{Loss_{FE}}$ of $\mathbf{Logits_{FE}}$ and $\mathbf{Lables_{FE}}$ from the given loss function;
            \State $\mathbf{Loss_A} = \mathbf{Loss_{FE} / \mathbf{A_{FE}}}$;
            \State $\mathbf{Loss_A}$ is appended to $\mathbf{Loss}$;
        \EndFor
        \State return $\mathbf{Loss, Logits}$;
    \end{algorithmic}
\end{algorithm}

For the backward pass, since FlashInfer does not support gradient computation, Loquetier falls back to the standard forward implementation backed by PyTorch’s Autograd when handling fine-tuning requests. This enables full gradient computation through efficient C++-based differentiation. For inference-only requests, Loquetier instead leverages FlashInfer’s batch-prefill and batch-decode kernels to maximize throughput and memory efficiency.

The forward computation flow for the attention layer is summarized in Algorithm~\ref{alg:comp-flow-attn}. First, the $\textbf{Q}$, $\textbf{K}$, and $\textbf{V}$ projections are computed jointly for all incoming requests. Attention outputs are then computed independently for each request type, concatenated, and passed through a shared output projection $\textbf{O}$.  
Algorithm~\ref{alg:comp-flow-model} outlines the forward method for the causal language model. For each incoming request, Loquetier computes output logits and, when applicable, the loss with respect to provided labels. Fine-tuning and evaluation requests include ground truth labels and thus return both logits and loss values, while prefilling and decoding requests return only logits.  
Loquetier enables joint forward pass of fine-tuning and evaluation requests within the same batch. Because the losses are tracked separately in Loquetier, this separation allows distinct gradient accumulation strategies for different fine-tuning tasks in parallel, without cross-interference. By summing losses across all fine-tuning requests, Loquetier produces a shared backward pass, enabling gradients from multiple fine-tuning jobs to be computed efficiently in a single backpropagation step.

We further extend the Transformers' Trainer \citep{wolf_transformers_2020} to support an interruptible fine-tuning process. Multiple trainers can now share the same computation flow in Loquetier, performing unified forward and backward propagation for fine-tuning different LoRA adapters concurrently. To ensure that each trainer only updates its corresponding parameters, we introduce \texttt{MixedLoRAModelForTrainer}, which applies parameter masking on top of a shared \texttt{MixedLoraModel} instance to achieve isolation.

%% file: figures/figure1.tex
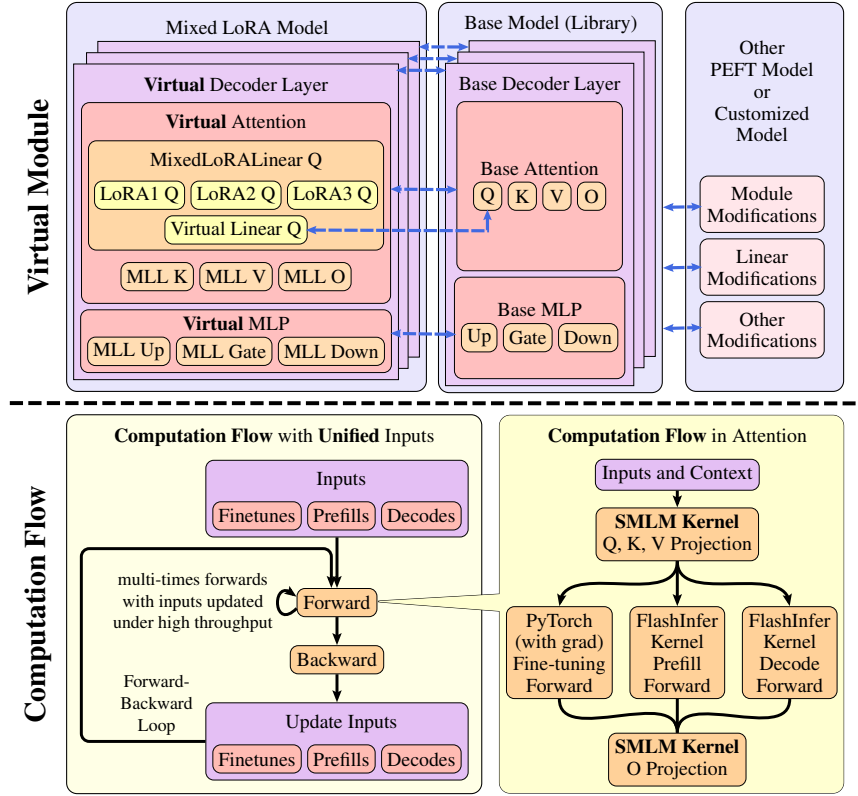
\begin{figure}
    \centering
    \begin{tikzpicture}[
        scale=0.75,
        every node/.style={transform shape},
        base/.style={draw, rectangle, rounded corners, fill=Purple!20!Blue!10},
        vmodule/.style={draw, rectangle, rounded corners, fill=Purple!20!Blue!10},
        vdlayer/.style={draw, rectangle, rounded corners=0, fill=DarkViolet!20, minimum width=5.7cm, minimum height=5.6cm},
        dlayer/.style={draw, rectangle, rounded corners=0, fill=DarkViolet!20, minimum width=3.3cm, minimum height=5.6cm},
        vattnmlp/.style={draw, rectangle, rounded corners, fill=Red!25, minimum width=5.2cm},
        attnmlp/.style={draw, rectangle, rounded corners, fill=Red!25, minimum width=2.7cm},
        linear/.style={draw, rectangle, rounded corners, fill=DarkOrange!30, minimum width=0.5cm, minimum height=0.5cm},
        vlinear/.style={draw, rectangle, rounded corners, fill=Yellow!30, minimum width=0.5cm, minimum height=0.5cm},
        modifies/.style={draw, rectangle, rounded corners, fill=Pink!40, minimum width=2.2cm, minimum height=1cm},
        nominh/.style={minimum height=0},
        nominw/.style={minimum width=0},
        bubble/.style={draw, rectangle callout, rounded corners, draw opacity=0, callout pointer width=0.3cm, minimum width=6.1cm, minimum height=6.7cm},
        bubbleptronly/.style={draw, rectangle callout, rounded corners, text opacity=0, fill opacity=0, draw opacity=1, callout pointer width=0.3cm, minimum width=6.1cm, minimum height=6.7cm},
        bubblefillonly/.style={draw, rectangle callout, rounded corners, fill=Yellow!18, text opacity=0, fill opacity=1, draw opacity=0, callout pointer width=0.3cm, minimum width=6.1cm, minimum height=6.7cm},
        inputs/.style={draw, rectangle, rounded corners, fill=DarkViolet!25, minimum width=0.5cm, minimum height=0.5cm},
        subinputs/.style={draw, rectangle, rounded corners, fill=DarkOrange!20!Red!30, minimum width=0.5cm, minimum height=0.5cm},
        compnode/.style={draw, rectangle, rounded corners, fill=DarkOrange!40, minimum width=0.5cm, minimum height=0.5cm},
        compflow/.style={arrows={-Stealth[length=6pt, inset=1pt]}, line width=1.5pt, color=Black},
        compflownoarrow/.style={line width=1.5pt, color=Black},
        vbproj/.style={dashed, dash pattern=on 5pt off 1pt, arrows={Stealth[length=4pt, inset=0.5pt]-Stealth[length=4pt, inset=0.5pt]}, line width=1.2pt, color=RoyalBlue},
        divline/.style={dashed, dash pattern=on 5pt off 2pt, line width=1.5pt, color=black},
    ]
    
    \node[base, minimum height=6.9cm] (mixed-model) {
        \begin{tikzpicture}[nominh]
            \node[rectangle, text opacity=1] (mm-title) {Mixed LoRA Model};
            \node[vdlayer, below right=0 and 0.2cm of mm-title.south, anchor=north] (vm-2) {};
            \node[vdlayer, below left=0.2cm and 0.2cm of vm-2.north, anchor=north] (vm-1) {};
            \node[vdlayer, below left=0.2cm and 0.2cm of vm-1.north, anchor=north] (vm-0) {
                \begin{tikzpicture}[nominh, minimum width=5.4cm]
                    \node[rectangle, text opacity=1] (vm-title) {\textbf{Virtual} Decoder Layer};
                    \node[vattnmlp, below=0 of vm-title] (v-attn) {
                        \begin{tikzpicture}[inner sep=1mm]
                            \node[rectangle, text opacity=1] (v-attn-title) {\textbf{Virtual} Attention};
                            \node[linear, below=1mm of v-attn-title, fill=DarkOrange!35] (mll-q) {
                                \begin{tikzpicture}[inner sep=1mm]
                                    \node[rectangle, text opacity=1] (mll-q-title) {MixedLoRALinear Q};
                                    \node[vlinear, below=1mm of mll-q-title] (l2q) {LoRA2 Q};
                                    \node[vlinear, left=1mm of l2q] (l1q) {LoRA1 Q};
                                    \node[vlinear, right=1mm of l2q] (l3q) {LoRA3 Q};
                                    \node[vlinear, below=1mm of l2q] (vq) {Virtual Linear Q};
                                \end{tikzpicture}
                            };
                            \node[below=1mm of mll-q] (v-attn-linears) {
                                \begin{tikzpicture}[inner sep=1mm]
                                    \node[linear] (mll-k) {MLL K};
                                    \node[linear, right=1mm of mll-k] (mll-v) {MLL V};
                                    \node[linear, right=1mm of mll-v] (mll-o) {MLL O};
                                \end{tikzpicture}
                            };
                        \end{tikzpicture}
                    };
                    \node[vattnmlp, below=1mm of v-attn] (v-mlp) {
                        \begin{tikzpicture}[inner sep=0]
                            \node[rectangle, text opacity=1] (v-mlp-title) {\textbf{Virtual} MLP};
                            \node[below=1mm of v-mlp-title] (v-mlp-linears) {
                                \begin{tikzpicture}[inner sep=1mm]
                                    \node[linear] (mll-up) {MLL Up};
                                    \node[linear, right=1mm of mll-up] (mll-gate) {MLL Gate};
                                    \node[linear, right=1mm of mll-gate] (mll-down) {MLL Down};
                                \end{tikzpicture}
                            };
                        \end{tikzpicture}};
                \end{tikzpicture}
            };
        \end{tikzpicture}
    };

    \node[base, right=0.2cm of mixed-model.north east, anchor=north west, minimum height=6.9cm] (base-model) {
        \begin{tikzpicture}[nominh]
            \node[rectangle, text opacity=1] (b-title) {Base Model (Library)};
            \node[dlayer, below right=0 and 0.2cm of b-title.south, anchor=north] (bm-2) {};
            \node[dlayer, below left=0.2cm and 0.2cm of bm-2.north, anchor=north] (bm-1) {};
            \node[dlayer, below left=0.2cm and 0.2cm of bm-1.north, anchor=north] (bm-0) {
                \begin{tikzpicture}[nominh, nominw]
                    \node[rectangle, text opacity=1] (bm-title) {Base Decoder Layer};
                    \node[attnmlp, below=0 of bm-title.south, minimum height=3cm] (b-attn) {
                        \begin{tikzpicture}[nominh, inner sep=0]
                            \node[rectangle, text opacity=1] (b-attn-title) {Base Attention};
                            \node[below=1mm of b-attn-title] (b-attn-linears) {
                                \begin{tikzpicture}[inner sep=1mm]
                                    \node[linear] (q) {Q};
                                    \node[linear, right=1mm of q] (k) {K};
                                    \node[linear, right=1mm of k] (v) {V};
                                    \node[linear, right=1mm of v] (o) {O};
                                \end{tikzpicture}
                            };
                        \end{tikzpicture}
                    };
                    \node[attnmlp, below=1mm of b-attn.south, minimum height=1.8cm] (b-mlp) {
                        \begin{tikzpicture}[nominh, inner sep=0]
                            \node[rectangle, text opacity=1] (b-mlp-title) {Base MLP};
                            \node[below=1mm of b-mlp-title] (b-mlp-linears) {
                                \begin{tikzpicture}[inner sep=1mm]
                                    \node[linear] (up) {Up};
                                    \node[linear, right=1mm of up] (gate) {Gate};
                                    \node[linear, right=1mm of gate] (down) {Down};
                                \end{tikzpicture}
                            };
                        \end{tikzpicture}};
                \end{tikzpicture}
            };
        \end{tikzpicture}
    };

    \node[base, right=0.4cm of base-model.north east, anchor=north west, align=center, minimum width=2.8cm, minimum height=6.9cm] (peft-model) {
        \begin{tikzpicture}[nominw, nominh, inner sep=1mm]
            \node[rectangle, text opacity=1, align=center] (title) {Other\\PEFT Model\\or\\Customized\\Model};
            \node[modifies, below=5mm of title] (mmod) {Module\\Modifications};
            \node[modifies, below=1mm of mmod] (lmod) {Linear\\Modifications};
            \node[modifies, below=1mm of lmod] (omod) {Other\\Modifications};
        \end{tikzpicture}
    };

    \node[base, fill=Yellow!10, below=0.4cm of mixed-model.south west, anchor=north west, minimum width=7cm, minimum height=6.7cm] (1f1b-comp-flow) {
        \begin{tikzpicture}[nominh, nominw]
            \node[rectangle, text opacity=1] (comp-flow-title) {\textbf{Computation Flow} with \textbf{Unified} Inputs};
            \node[rectangle, below=0 of comp-flow-title] (comp-flow) {
                \begin{tikzpicture}[text opacity=1, fill opacity=1, draw opacity=1]
                    \node[inputs] (inputs) {
                        \begin{tikzpicture}[inner sep=1mm]
                            \node[rectangle, text opacity=1] (inputs-title) {Inputs};
                            \node[subinputs, below=1mm of inputs-title] (prefills) {Prefills};
                            \node[subinputs, left=1mm of prefills] (finetunes) {Finetunes};
                            \node[subinputs, right=1mm of prefills] (decodes) {Decodes};
                        \end{tikzpicture}
                    };
                    \node[compnode, below=0.9cm of inputs] (forward) {Forward};
                    \node[compnode, below=0.5cm of forward] (backward) {Backward};
                    \node[inputs, below=0.5cm of backward] (new-inputs) {
                        \begin{tikzpicture}[inner sep=1mm]
                            \node[rectangle, text opacity=1] (inputs-title) {Update Inputs};
                            \node[subinputs, below=1mm of inputs-title] (prefills) {Prefills};
                            \node[subinputs, left=1mm of prefills] (finetunes) {Finetunes};
                            \node[subinputs, right=1mm of prefills] (decodes) {Decodes};
                        \end{tikzpicture}
                    };
                    \node[rectangle, text opacity=1, align=center, above left=0 and 0.1cm of new-inputs.west] (loop) {\small{Forward-}\\\small{Backward}\\\small{Loop}};
        
                    \draw[compflow] (inputs) -- (forward);
                    \draw[compflow] (forward) -- (backward);
                    \draw[compflow] (backward) -- (new-inputs);
        
                    \draw[compflow] ($(forward.south west)!0.3!(forward.north west)$) to[out=225, in=135, loop, looseness=8] ($(forward.south west)!0.7!(forward.north west)$);
                    \node[rectangle, text opacity=1, align=center, left=0.3cm of forward.west] (forward-loop) {\small{multi-times forwards}\\\small{with inputs updated}\\\small{under high throughput}};
                    
                    \coordinate (west-of-new-inputs) at ($(new-inputs.west) - (2.2cm,0)$);
                    \coordinate (west-north-of-forward) at ($(forward.north) - (0.1cm,-0.7cm)$);
                    \coordinate (west-north-at-forward) at ($(forward.north) - (0.1cm,0)$);
                    \draw[compflow] (new-inputs.west) -- (west-of-new-inputs) -- ($(west-of-new-inputs |- west-north-of-forward)$) -- (west-north-of-forward) -- (west-north-at-forward);
                \end{tikzpicture}
            };
        \end{tikzpicture}
    };

    \node[bubbleptronly, right=0.3cm of 1f1b-comp-flow, callout absolute pointer={($(1f1b-comp-flow.east) + (-1.95cm, 0.05cm)$)}] (attn-comp-flow-ptr2) {};
    \node[bubblefillonly, right=0.3cm of 1f1b-comp-flow, callout absolute pointer={($(1f1b-comp-flow.east) + (-1.95cm, 0.05cm)$)}] (attn-comp-flow-fill2) {};
    \node[bubble, right=0.3cm of 1f1b-comp-flow, callout absolute pointer={($(mixed-model.east) + (-0.7cm, 0)$)}] (attn-comp-flow) {
        \begin{tikzpicture}[nominh, nominw, draw opacity=1]
            \node[rectangle, text opacity=1] (comp-flow-title) {\textbf{Computation Flow} in Attention};
            \node[inputs, below=1mm of comp-flow-title] (inputs) {Inputs and Context};
            \node[compnode, below=0.3cm of inputs, align=center] (qkv-proj) {\textbf{SMLM Kernel}\\Q, K, V Projection};
            \node[compnode, below=0.8cm of qkv-proj, align=center] (pf-forward) {FlashInfer\\Kernel\\Prefill\\Forward};
            \node[compnode, left=0.3cm of pf-forward, align=center] (ft-forward) {PyTorch\\(with grad)\\Fine-tuning\\Forward};
            \node[compnode, right=0.3cm of pf-forward, align=center] (dc-forward) {FlashInfer\\Kernel\\Decode\\Forward};
            \node[compnode, below=0.6cm of pf-forward, align=center] (o-proj) {\textbf{SMLM Kernel}\\O Projection};

            \draw[compflow] (inputs) -- (qkv-proj);
            \draw[compflow] (qkv-proj) -- (pf-forward);
            \draw[compflow] (qkv-proj) to[out=270, in=90] (ft-forward);
            \draw[compflow] (qkv-proj) to[out=270, in=90] (dc-forward);
            \draw[compflow] (pf-forward) to[out=270, in=90] (o-proj);
            \draw[compflownoarrow] (ft-forward.south) to[out=270, in=90] (o-proj);
            \draw[compflownoarrow] (dc-forward) to[out=270, in=90] (o-proj);
        \end{tikzpicture}
    };
    
    \node[rectangle, left=0.5cm of mixed-model, text opacity=1, align=center, scale=1.5, rotate=90, anchor=center, transform shape] (vm-part-title) {\textbf{Virtual Module}};
    \node[rectangle, left=0.5cm of 1f1b-comp-flow, text opacity=1, align=center, scale=1.5, rotate=90, anchor=center, transform shape] (cf-part-title) {\textbf{Computation Flow}};
    
    \coordinate (div-left) at ($(mixed-model.south west)!0.5!(1f1b-comp-flow.north west) - (1cm,0)$);
    \coordinate (div-right) at ($(peft-model.south east)!0.5!(attn-comp-flow.north east) + (0.2cm,0)$);
    
    \draw[divline] (div-left) -- (div-right);

    \coordinate (base-dlayer-2) at ($(base-model.north west)!0.115!(base-model.south west)$);
    \coordinate (base-dlayer-1) at ($(base-model.north west)!0.145!(base-model.south west)$);
    \coordinate (base-dlayer-0) at ($(base-model.north west)!0.175!(base-model.south west)$);
    \coordinate (base-attn) at ($(base-model.north west)!0.48!(base-model.south west)$);
    \coordinate (base-mlp) at ($(base-model.north west)!0.85!(base-model.south west)$);
    \coordinate (base-q) at ($(base-model.north west)!0.53!(base-model.south west)$);
    \coordinate (base-q-south) at ($(base-model.north west)!0.585!(base-model.south west)$);
    \coordinate (mixed-dlayer-2) at ($(mixed-model.north east)!0.115!(mixed-model.south east)$);
    \coordinate (mixed-dlayer-1) at ($(mixed-model.north east)!0.145!(mixed-model.south east)$);
    \coordinate (mixed-dlayer-0) at ($(mixed-model.north east)!0.175!(mixed-model.south east)$);
    \coordinate (mixed-attn) at ($(mixed-model.north east)!0.48!(mixed-model.south east)$);
    \coordinate (mixed-mlp) at ($(mixed-model.north east)!0.85!(mixed-model.south east)$);
    \coordinate (mixed-q) at ($(mixed-model.north east)!0.585!(mixed-model.south east)$);
    
    \draw[vbproj] ($(base-dlayer-2) + (0.55cm,0)$) -- ($(mixed-dlayer-2) - (0.15cm,0)$);
    \draw[vbproj] ($(base-dlayer-1) + (0.35cm,0)$) -- ($(mixed-dlayer-1) - (0.35cm,0)$);
    \draw[vbproj] ($(base-dlayer-0) + (0.15cm,0)$) -- ($(mixed-dlayer-0) - (0.55cm,0)$);
    \draw[vbproj] ($(base-attn) + (0.35cm,0)$) -- ($(mixed-attn) - (0.65cm,0)$);
    \draw[vbproj] ($(base-mlp) + (0.325cm,0)$) -- ($(mixed-mlp) - (0.65cm,0)$);
    \draw[vbproj] ($(base-q) + (0.875cm,0)$) -- ($(base-q-south) + (0.875cm,0)$) -- ($(mixed-q) - (2.15cm,0)$);
    
    \coordinate (base-mod-0) at ($(base-model.north east)!0.525!(base-model.south east)$);
    \coordinate (base-mod-1) at ($(base-model.north east)!0.68!(base-model.south east)$);
    \coordinate (base-mod-2) at ($(base-model.north east)!0.835!(base-model.south east)$);
    \coordinate (peft-mod-0) at ($(peft-model.north west)!0.525!(peft-model.south west)$);
    \coordinate (peft-mod-1) at ($(peft-model.north west)!0.68!(peft-model.south west)$);
    \coordinate (peft-mod-2) at ($(peft-model.north west)!0.835!(peft-model.south west)$);
    
    \draw[vbproj] (base-mod-0) -- ($(peft-mod-0) + (0.3cm,0)$);
    \draw[vbproj] (base-mod-1) -- ($(peft-mod-1) + (0.3cm,0)$);
    \draw[vbproj] (base-mod-2) -- ($(peft-mod-2) + (0.3cm,0)$);
    
    \end{tikzpicture}
    \caption{The framework diagram of Loquetier.}
    \label{fig:framework-struct}
\end{figure}

%% file: chapters/chapter4.tex
\begin{table}
    \caption{Comparison on model loading between Loquetier, PEFT, S-LoRA and FlexLLM. Metrics include time to load (Time) and additional storage footprint (Storage).}
    \label{tab:loading}
    \centering
    \begin{tabular}{lllllll}
        \toprule
        & \multicolumn{2}{c}{Base Model} & \multicolumn{2}{c}{LoRA Model} &
        \multicolumn{2}{c}{Total}                                      \\
        \cmidrule(r){2-3} \cmidrule(r){4-5} \cmidrule(r){6-7}
        Framework or System &
        Time & Storage & Time & Storage & Time & Storage               \\
        \midrule
        Loquetier &
        2.927 s & 0 B & 2.409 s & 0 B & 5.336 s & 0 B                  \\
        PEFT &
        2.877 s & 0 B & 1.914 s & 0 B & 4.791 s & 0 B                  \\
        S-LoRA &
        33.037 s & 0 B & 0.948 s & 0 B & 33.985 s & 0 B                  \\
        FlexLLM &
        37.933 s & 14.96 GB & 0.924 s & 40.04 MB & 38.857 s & 15.00 GB \\
        \bottomrule
    \end{tabular}
\end{table}

\section{Experiments}

The Loquetier framework expects to enable fine-tuning and reasoning across multiple LoRA models. Based on the objective, we design three experiments to evaluate the performance of Loquetier in these scenarios: inference-only, fine-tuning-only, and unified fine-tuning and inference.
In addition, we perform two experiments to evaluate the performance of Loquetier in simulated real-world environments: a shorter, approximate simulation for rapidly testing different load conditions across different times of the day, and a 120-minute precise simulation using data extracted from real-world scenarios for a more demanding and realistic stress test.
To further validate the effectiveness of our design, we conduct a micro-experiment to evaluate the efficiency of the model loading process.

\subsection{Evaluation settings}
\label{sec:exp-settings}

\textbf{Baselines}. FlexLLM is an advanced co-serving system for LLM serving and parameter-efficient fine-tuning. We deployed docker images as its runtime environment following the guidelines. 
S-LoRA is designed specifically for large-scale LoRA inference. Therefore, we combine S-LoRA with PEFT as another baseline, in which PEFT handles the fine-tuning task. 
We use HuggingFace Transformers with PEFT as the most basic baseline.

\input{figures/figure2}

\textbf{Models}.
We use the Llama3-8B model as the base model. The LoRA adapter is obtained by training on the Alpaca dataset. For the fine-tuning task, we use the same LoRA configuration as the inference LoRA adapter and initialize the weights from the Gaussian distribution.

\textbf{Datasets}. We use the ShareGPT dataset as input for the inference task, and the Alpaca and GSM8K datasets as input for the fine-tuning task.
BurstGPT \citep{wang_burstgpt_2025} is an LLM service workload dataset comprising over ten million traces collected from Azure OpenAI GPT services. Data spanning more than 60 days is segmented into 20-minute slices according to its provided partitioning scheme.

\textbf{Hardware}.
We test the inference-only tasks on a server with NVIDIA A6000 48G GPUs. We test the fine-tuning-only tasks and the unified fine-tuning and inference tasks on servers with NVIDIA H800 80G GPUs. Each test process has at least 128G of host memory available.

\input{figures/figure3}

\textbf{Metrics and Tasks}.
The metrics used for the evaluation are listed in Appendix~\ref{sec:metrics}.
The inference-only tasks run inference tasks at different request arrival rates where inference needs to be as fast as possible to achieve the Service Level Objective (SLO). The fine-tuning-only tasks need to run the training task for a specified number of epochs to measure its efficiency in processing tokens. The  unified fine-tuning and inference tasks need to run the training task along with the inference task, weighing the efficiency of fine-tuning tokens against the SLO of the inference request. 
The real-world workload simulation experiment samples six time periods from the BurstGPT dataset, comprising one low-load, two medium-load, and three high-load intervals. Each slice was categorized into one of three load tiers based on its request rate: low-load periods for average RPS $<1$; medium-load for $1$\textasciitilde$1.75$; and high-load for $>1.75$ (which may include transient spikes exceeding RPS 10). Here, RPS denotes requests per second.
The detailed configurations can be viewed in Appendix~\ref{sec:settings}.

\subsection{Evaluation results}

\input{figures/figure4}

\textbf{Model Loading}.\label{sec:model-loading} The loading speed and additional loading storage overhead are shown in Table~\ref{tab:loading}. Compared to PEFT, Loquetier requires creating Virtualized Modules and applying scaling to each LoRA linear when loading LoRA models, resulting in a slight slowdown in this part of the loading. FlexLLM needs to transform and cache the model weights, which leads to a significant additional storage footprint. Even with cached transformed models, FlexLLM's loading is still very slow due to the need to read small weight files.

\textbf{Inference}.
Figure~\ref{fig:test-infer} shows the test results of the inference task. Loquetier maintains the highest SLO attainment at different request arrival rates. As the request rate increases, Loquetier's decoding speed gradually increases. Until at 3 RPS, the decoding speed no longer increases, indicating that the GPU memory access bottleneck has been hit. As the request rate continues to increase, some requests begin to fail to reach their SLOs due to the inability to achieve faster inference on the current GPU.

FlexLLM's maximum decoding speed is lower than Loquetier's, causing its SLO attainment rate to start dropping earlier and fall off a cliff at higher request arrival rates. In addition, in conjunction with the findings in Section~\ref{sec:model-loading}, FlexLLM's lazy loading mechanism prevents it from handling some of the earliest arriving requests under SLO. Forcing early loading of the model weights improves SLO attainment to or near 100\% at 1-2 RPS, but the improvement is very limited for higher request arrival rates due to the limitations of its highest decoding speed.
FlexLLM is unable to apply LoRA to all 7 modules, causing it to fail under the corresponding experiments. When loading multiple LoRA models, FlexLLM is trapped in a dead loop for more than 10 minutes, missing SLOs for all requests. After a longer wait, FlexLLM cannot get out of the trap still, and therefore is marked as a failure.

Transformers' batch strategy of padding different inputs to the same length makes its GPU memory footprint greatly affected by the batch size, making it very easy to trigger the error "CUDA out of memory". The processing speed of PEFT is constrained by the batch size to avoid exceeding the GPU memory. In multiple LoRA inference tasks, PEFT can only apply LoRAs in a serial for different configurations of inputs, making the inference speed further degraded. PEFT's SLO attainment rate is unacceptable even under 1 RPS.

\textbf{Fine-tuning}. The test results for the fine-tuning task are shown in Figure~\ref{fig:test-ft}. The shorter total training time for Loquetier is due to its faster evaluation. Loquetier's fine-tuning is slightly slower than that of PEFT, mainly because of the independent computational calls from the LoRA linears during backward propagation. FlexLLM encounters an unsupported operation error during its peft backward propagation, indicating its inability to complete the experiments. The results show that Loquetier leads to almost no loss of fine-tuning efficiency.

\textbf{Unified Fine-tuning and Inference}. We test a combination of different configurations of fine-tuning and inference tasks, and the results are shown in Figure~\ref{fig:test-unified}. Loquetier is able to provide an average of about 40\% fine-tuning efficiency over three different request arrival rates while maintaining similar SLOs as in the inference-only tasks. PEFT's inference efficiency is too low, resulting in over 90\% of the inference tasks timing out before they even begin. PEFT's fine-tuning tasks only drop about 20\% efficiency, but this is due to the fact that PEFT has almost no computational overhead on the inference tasks, allowing the vast majority of the computing resources to still be used by the fine-tuning tasks.
FlexLLM fails to complete the fine-tuning task, so it is not available for unified tests.

\textbf{Mutable Capacity Allocation Simulation}.
In order to evaluate performance facing dynamic loads in real-world scenarios, we design an inference subtask with dynamic input throughput. As shown in Figure~\ref{fig:test-mutable}, Loquetier is able to adaptively adjust the efficiency of both fine-tuning and inference tasks under dynamic loads in the mutable unified task. The fine-tuning task makes concessions for the inference task to ensure the quality of service when request throughput increases, and adjusts back the efficiency by itself when throughput decreases.

\input{figures/figure5}
\input{figures/figure6}
\textbf{Simulated Real-world Workload}.
To better simulate real-world serving conditions, we construct a composite workload using slices from the BurstGPT dataset, as mentioned in Section~\ref{sec:exp-settings}. Each slice contains request arrival times, input lengths, and output lengths. In our simulation, we fully utilize the request arrival times and reference the input length data. For an overview of sampling data and preference adjustments, see Appendix~\ref{sec:simr}.
As shown in Figure~\ref{fig:test-burstgpt}, Loquetier demonstrates strong adaptability to real-world workloads, aligning closely with trends evaluated in the earlier simulation experiments. The final SLO for the entire experiment reaches 92.37\%. All requests that failed to meet service metrics occur during transient workload spikes under high-load conditions (RPS $>$ 5), which exceeded the load capacity of the hardware. In all other periods, Loquetier consistently achieves the defined SLO.

%% file: figures/figure2.tex
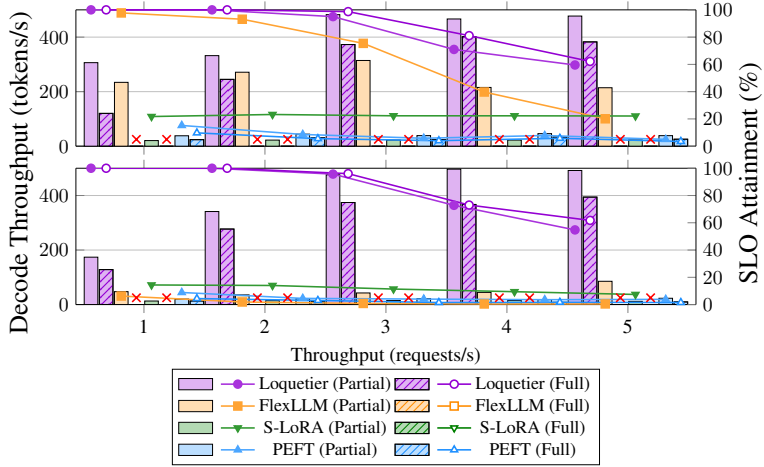
\begin{figure}
    \centering
    \begin{tikzpicture}[scale=0.75]
        \def \plotmainheight {8cm}
        \def \labeloffsetx {0.2cm}
        \def \labeloffsety {-0.2cm}
        \begin{axis}[
            name=plot431,
            width=0.9\textwidth,
            height=\plotmainheight*0.5,
            ybar,
            axis y line*=left,
            bar width=0.9,
            bar shift=0,
            ymajorgrids=true,
            ymin=0, ymax=500,
            xtick={3.5,11.5,19.5,27.5,35.5},
            xticklabels=\empty,
            xmin=-1, xmax=40,
            legend=none
        ]
        
        \addplot[fill=DarkViolet!30] coordinates {(0,306.33) (8,332.1) (16,482.43) (24,466.56) (32,477.27)};
        \addplot[fill=DarkViolet!30, postaction={pattern=north east lines, pattern color=DarkViolet}] coordinates {(1,120.28) (9,245.03) (17,372.81) (25,401.46) (33,382.47)};
        \addplot[fill=DarkOrange!30] coordinates {(2,234.12) (10,271.26) (18,314.37) (26,215.47) (34,214.37)};
        \addplot[fill=DarkOrange!30, postaction={pattern=north east lines, pattern color=DarkOrange}] coordinates {(3,0) (11,0) (19,0) (27,0) (35,0)};
        \addplot[fill=Green!30] coordinates {(4,20.58) (12,22.31) (20,22.5) (28,22.47) (36,22.86)};
        \addplot[fill=Green!30, postaction={pattern=north east lines, pattern color=Green}] coordinates {(5,0) (13,0) (21,0) (29,0) (37,0)};
        \addplot[fill=DodgerBlue!30] coordinates {(6,38.18) (14,44.09) (22,38.74) (30,45.96) (38,38.63)};
        \addplot[fill=DodgerBlue!30, postaction={pattern=north east lines, pattern color=DodgerBlue}] coordinates {(7,23.89) (15,30.59) (23,24.49) (31,33.19) (39,25.52)};
        \end{axis}
        
        \begin{axis}[
            width=0.9\textwidth,
            height=\plotmainheight*0.5,
            axis y line*=right,
            ymin=0, ymax=100,
            xtick={3.5,11.5,19.5,27.5,35.5},
            xmin=-1, xmax=40,
            legend=none,
            hide x axis
          ]
        
        \addplot[mark=*, DarkViolet!80, thick] coordinates {(0,100) (8,100) (16,95.04) (24,70.97) (32,59.55)};
        \addplot[mark=*, mark options={fill=white, draw=DarkViolet}, DarkViolet, thick] coordinates {(1,100) (9,100) (17,98.71) (25,81.09) (33,62.18)};
        \addplot[mark=square*, DarkOrange!80, thick] coordinates {(2,97.75) (10,93.12) (18,75.42) (26,39.72) (34,20.1)};
        \addplot[only marks, mark=x, mark options={scale=1.5}, Red, thick] coordinates {(3,5) (11,5) (19,5) (27,5) (35,5)};
        \addplot[mark=triangle*, mark options={rotate=180}, Green!80, thick] coordinates {(4,21.75) (12,23.25) (20,22.29) (28,22.28) (36,22.13)};
        \addplot[only marks, mark=x, mark options={scale=1.5}, Red, thick] coordinates {(5,5) (13,5) (21,5) (29,5) (37,5)};
        \addplot[mark=triangle*, DodgerBlue!80, thick] coordinates {(6,15.2) (14,8.59) (22,5.81) (30,7.83) (38,4.99)};
        \addplot[mark=triangle*, mark options={fill=white, draw=DodgerBlue}, DodgerBlue, thick] coordinates {(7,9.76) (15,5.54) (23,4.03) (31,5.79) (39,3.47)};
        \end{axis}
        
        \begin{axis}[
            name=plot432,
            at={(plot431.below south west)},
            anchor=north west,
            width=0.9\textwidth,
            height=\plotmainheight*0.5,
            ybar,
            axis y line*=left,
            bar width=0.9,
            bar shift=0,
            ymajorgrids=true,
            ymin=0, ymax=500,
            xtick={3.5,11.5,19.5,27.5,35.5},
            xticklabels={1,2,3,4,5},
            xlabel=Throughput (requests/s),
            xmin=-1, xmax=40,
            legend=none
        ]
        
        \addplot[fill=DarkViolet!30] coordinates {(0,173.84) (8,341.25) (16,476.05) (24,496.61) (32,491.76)};
        \addplot[fill=DarkViolet!30, postaction={pattern=north east lines, pattern color=DarkViolet}] coordinates {(1,128.25) (9,276.99) (17,374.34) (25,367.15) (33,394.17)};
        \addplot[fill=DarkOrange!30] coordinates {(2,47.33) (10,35.46) (18,42.59) (26,44.68) (34,85.38)};
        \addplot[fill=DarkOrange!30, postaction={pattern=north east lines, pattern color=DarkOrange}] coordinates {(3,0) (11,0) (19,0) (27,0) (35,0)};
        \addplot[fill=Green!30] coordinates {(4,13.02) (12,14.21) (20,14.67) (28,15.01) (36,11.79)};
        \addplot[fill=Green!30, postaction={pattern=north east lines, pattern color=Green}] coordinates {(5,0) (13,0) (21,0) (29,0) (37,0)};
        \addplot[fill=DodgerBlue!30] coordinates {(6,20.09) (14,21.47) (22,22.09) (30,18.28) (38,23.09)};
        \addplot[fill=DodgerBlue!30, postaction={pattern=north east lines, pattern color=DodgerBlue}] coordinates {(7,13.21) (15,9.98) (23,6.83) (31,8.02) (39,9.67)};
        \end{axis}
        
        \begin{axis}[
            at={(plot431.below south west)},
            anchor=north west,
            width=0.9\textwidth,
            height=\plotmainheight*0.5,
            axis y line*=right,
            ymin=0, ymax=100,
            xtick={3.5,11.5,19.5,27.5,35.5},
            xmin=-1, xmax=40,
            legend=none,
            hide x axis
          ]
        
        \addplot[mark=*, DarkViolet!80, thick] coordinates {(0,100) (8,100) (16,95.71) (24,72.81) (32,54.77)};
        \addplot[mark=*, mark options={fill=white, draw=DarkViolet}, DarkViolet, thick] coordinates {(1,100) (9,100) (17,96.04) (25,73) (33,61.75)};
        \addplot[mark=square*, DarkOrange!80, thick] coordinates {(2,6.25) (10,1.88) (18,0.83) (26,0.44) (34,0.47)};
        \addplot[only marks, mark=x, mark options={scale=1.5}, Red, thick] coordinates {(3,5) (11,5) (19,5) (27,5) (35,5)};
        \addplot[mark=triangle*, mark options={rotate=180}, Green!80, thick] coordinates {(4,14.38) (12,14) (20,11.38) (28,9.44) (36,7.42)};
        \addplot[only marks, mark=x, mark options={scale=1.5}, Red, thick] coordinates {(5,5) (13,5) (21,5) (29,5) (37,5)};
        \addplot[mark=triangle*, DodgerBlue!80, thick] coordinates {(6,8.86) (14,4.53) (22,4.02) (30,3.58) (38,3.82)};
        \addplot[mark=triangle*, mark options={fill=white, draw=DodgerBlue}, DodgerBlue, thick] coordinates {(7,4.4) (15,3.01) (23,1.5) (31,1.69) (39,1.5)};
        \end{axis}
        
        \path (current bounding box.north) -- (current bounding box.south)
            coordinate[midway] (vertical_center);
        
        \node[rotate=90, anchor=center] at ($(current bounding box.west) - (\labeloffsetx,\labeloffsety)$)
            {Decode Throughput (tokens/s)};
        
        \node[rotate=90, anchor=center] at ($(current bounding box.east) - (-\labeloffsetx,\labeloffsety)$)
            {SLO Attainment (\%)};
        
        \begin{axis}[
            at={(plot432.below south west)},
            anchor=north west,
            yshift=-0.8cm,
            width=0.9\textwidth,
            height=\plotmainheight*0.2,
            ymin=0, ymax=1,
            xmin=0, xmax=1,
            axis line style={draw=none},
            xtick=\empty,
            ytick=\empty,
            legend style={
                nodes={scale=0.9, transform shape},
                legend columns=4,
                legend pos=north west,
                at={(0.5, 0.7)},
                anchor=center
            }
        ]
        
        \addlegendimage{area legend, fill=DarkViolet!30}
        \addlegendentry{}
        \addlegendimage{DarkViolet!80, mark=*, thick}
        \addlegendentry{Loquetier (Partial)}
        \addlegendimage{area legend, fill=DarkViolet!30, postaction={pattern=north east lines, pattern color=DarkViolet}}
        \addlegendentry{}
        \addlegendimage{DarkViolet, mark=*, mark options={fill=white, draw=DarkViolet}, thick}
        \addlegendentry{Loquetier (Full)}
        \addlegendimage{area legend, fill=DarkOrange!30}
        \addlegendentry{}
        \addlegendimage{DarkOrange!80, mark=square*, thick}
        \addlegendentry{FlexLLM (Partial)}
        \addlegendimage{area legend, fill=DarkOrange!30, postaction={pattern=north east lines, pattern color=DarkOrange}}
        \addlegendentry{}
        \addlegendimage{DarkOrange, mark=square*, mark options={fill=white, draw=DarkOrange}, thick}
        \addlegendentry{FlexLLM (Full)}
        \addlegendimage{area legend, fill=Green!30}
        \addlegendentry{}
        \addlegendimage{Green!80, mark=triangle*, mark options={rotate=180}, thick}
        \addlegendentry{S-LoRA (Partial)}
        \addlegendimage{area legend, fill=Green!30, postaction={pattern=north east lines, pattern color=Green}}
        \addlegendentry{}
        \addlegendimage{Green, mark=triangle*, mark options={fill=white, draw=Green, rotate=180}, thick}
        \addlegendentry{S-LoRA (Full)}
        \addlegendimage{area legend, fill=DodgerBlue!30}
        \addlegendentry{}
        \addlegendimage{DodgerBlue!80, mark=triangle*, thick}
        \addlegendentry{PEFT (Partial)}
        \addlegendimage{area legend, fill=DodgerBlue!30, postaction={pattern=north east lines, pattern color=DodgerBlue}}
        \addlegendentry{}
        \addlegendimage{DodgerBlue, mark=triangle*, mark options={fill=white, draw=DodgerBlue}, thick}
        \addlegendentry{PEFT (Full)}
        
        \end{axis}
    
    \end{tikzpicture}
    \caption{Comparison of the performance of Loquetier, FlexLLM, S-LoRA and PEFT in inference tasks. The upper is single LoRA model inference and the lower part is multiple LoRA model inference. Partial means that only 3 modules are enabled for FlexLLM including up, gate, and down. For detailed information on S-LoRA, please refer to the Appendix~\ref{sec:slora}. Full means that all 7 modules are enabled, including q, k, v, o, up, gate, and down. $\times$ indicates that the results were not obtained: FlexLLM does not support enabling LoRA modules for linear layers other than up, gate, and down; FlexLLM cycles through loading LoRA models during multi-LoRA inference.}
    \label{fig:test-infer}
\end{figure}

 

%% file: figures/figure3.tex
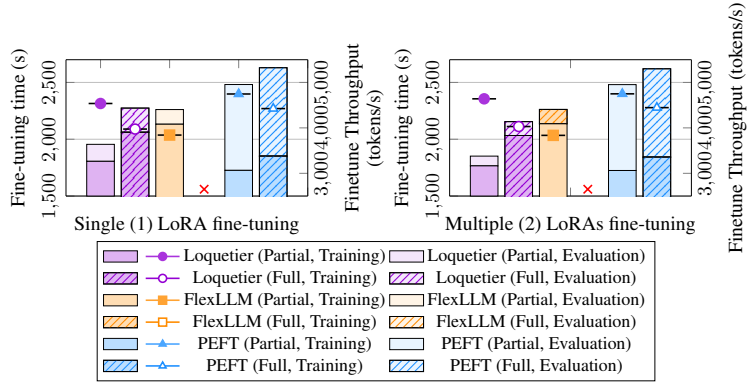
\begin{figure}
    \centering
    \begin{tikzpicture}[scale=0.75]
        \def \subplotsep {0.18\textwidth}
        \def \plotmainheight {8cm}
        \def \labeloffsetx {0.2cm}
        \def \labeloffsety {-0.2cm}
        \begin{axis}[
            name=plot441,
            width=0.42\textwidth,
            height=\plotmainheight*0.5,
            ybar stacked,
            axis y line*=left,
            bar width=0.8,
            bar shift=0,
            ymajorgrids=true,
            ymin=1500, ymax=2700,
            ylabel=Fine-tuning time (s),
            y tick label style={rotate=90},
            xticklabels=\empty,
            xmin=-1, xmax=6,
            xlabel=Single (1) LoRA fine-tuning,
            legend=none
        ]
        
        \addplot+[fill=DarkViolet!30, draw=black] coordinates {(0,1805.75) (1,0) (2,0) (3,0) (4,0) (5,0)};
        \addplot+[fill=DarkViolet!10, draw=black] coordinates {(0,149.39) (1,0) (2,0) (3,0) (4,0) (5,0)};
        \addplot+[fill=DarkViolet!30, draw=black, postaction={pattern=north east lines, pattern color=DarkViolet}] coordinates {(0,0) (1,2061.48) (2,0) (3,0) (4,0) (5,0)};
        \addplot+[fill=DarkViolet!10, draw=black, postaction={pattern=north east lines, pattern color=DarkViolet}] coordinates {(0,0) (1,212.86) (2,0) (3,0) (4,0) (5,0)};
        \addplot+[fill=DarkOrange!30, draw=black] coordinates {(0,0) (1,0) (2,2132.35) (3,0) (4,0) (5,0)};
        \addplot+[fill=DarkOrange!10, draw=black] coordinates {(0,0) (1,0) (2,128.93) (3,0) (4,0) (5,0)};
        \addplot+[fill=DarkOrange!30, draw=black, postaction={pattern=north east lines, pattern color=DarkOrange}] coordinates {(0,0) (1,0) (2,0) (3,0) (4,0) (5,0)};
        \addplot+[fill=DodgerBlue!30, draw=black] coordinates {(0,0) (1,0) (2,0) (3,0) (4,1726.17) (5,0)};
        \addplot+[fill=DodgerBlue!10, draw=black] coordinates {(0,0) (1,0) (2,0) (3,0) (4,755.46) (5,0)};
        \addplot+[fill=DodgerBlue!30, draw=black, postaction={pattern=north east lines, pattern color=DodgerBlue}] coordinates {(0,0) (1,0) (2,0) (3,0) (4,0) (5,1851.45)};
        \addplot+[fill=DodgerBlue!10, draw=black, postaction={pattern=north east lines, pattern color=DodgerBlue}] coordinates {(0,0) (1,0) (2,0) (3,0) (4,0) (5,778.09)};
        \end{axis}
        
        \begin{axis}[
            width=0.42\textwidth,
            height=\plotmainheight*0.5,
            axis y line*=right,
            ymin=2500, ymax=5500,
            ylabel style={align=center},
            ylabel={Finetune Throughput\\(tokens/s)},
            y tick label style={rotate=90},
            xmin=-1, xmax=6,
            legend=none,
            hide x axis
          ]
        
        \addplot[mark=-, mark options={scale=3}, black, thick] coordinates {(0,4536.62)};
        \addplot[mark=-, mark options={scale=3}, black, thick] coordinates {(1,3973.85)};
        \addplot[mark=-, mark options={scale=3}, black, thick] coordinates {(2,3841.77)};
        \addplot[mark=-, mark options={scale=3}, black, thick] coordinates {(4,4745.76)};
        \addplot[mark=-, mark options={scale=3}, black, thick] coordinates {(5,4424.65)};
        
        \addplot[mark=*, mark options={scale=1.25}, DarkViolet!80, thick] coordinates {(0,4536.62)};
        \addplot[mark=*, mark options={scale=1.25, fill=white, draw=DarkViolet}, DarkViolet, thick] coordinates {(1,3973.85)};
        \addplot[mark=square*, mark options={scale=1.25}, DarkOrange!80, thick] coordinates {(2,3841.77)};
        \addplot[only marks, mark=x, mark options={scale=1.5}, Red, thick] coordinates {(3,2650)};
        \addplot[mark=triangle*, mark options={scale=1.25}, DodgerBlue!80, thick] coordinates {(4,4745.76)};
        \addplot[mark=triangle*, mark options={scale=1.25, fill=white, draw=DodgerBlue}, DodgerBlue, thick] coordinates {(5,4424.65)};
        \end{axis}
        
        \begin{axis}[
            name=plot442,
            at={(plot441.east)},
            anchor=west,
            xshift=\subplotsep,
            width=0.42\textwidth,
            height=\plotmainheight*0.5,
            ybar stacked,
            axis y line*=left,
            bar width=0.8,
            bar shift=0,
            ymajorgrids=true,
            ymin=1500, ymax=2700,
            ylabel=Fine-tuning time (s),
            y tick label style={rotate=90},
            xtick={2.5},
            xticklabels=\empty,
            xmin=-1, xmax=6,
            xlabel=Multiple (2) LoRAs fine-tuning,
            legend=none
        ]
        
        \addplot+[fill=DarkViolet!30, draw=black] coordinates {(0,1765.75) (1,0) (2,0) (3,0) (4,0) (5,0)};
        \addplot+[fill=DarkViolet!10, draw=black] coordinates {(0,85.7) (1,0) (2,0) (3,0) (4,0) (5,0)};
        \addplot+[fill=DarkViolet!30, draw=black, postaction={pattern=north east lines, pattern color=DarkViolet}] coordinates {(0,0) (1,2032.92) (2,0) (3,0) (4,0) (5,0)};
        \addplot+[fill=DarkViolet!10, draw=black, postaction={pattern=north east lines, pattern color=DarkViolet}] coordinates {(0,0) (1,122.11) (2,0) (3,0) (4,0) (5,0)};
        \addplot+[fill=DarkOrange!30, draw=black] coordinates {(0,0) (1,0) (2,2135.99) (3,0) (4,0) (5,0)};
        \addplot+[fill=DarkOrange!30, draw=black, postaction={pattern=north east lines, pattern color=DarkOrange}] coordinates {(0,0) (1,0) (2,126.47) (3,0) (4,0) (5,0)};
        \addplot+[fill=DodgerBlue!30, draw=black] coordinates {(0,0) (1,0) (2,0) (3,0) (4,1724.94) (5,0)};
        \addplot+[fill=DodgerBlue!10, draw=black] coordinates {(0,0) (1,0) (2,0) (3,0) (4,755.84) (5,0)};
        \addplot+[fill=DodgerBlue!30, draw=black, postaction={pattern=north east lines, pattern color=DodgerBlue}] coordinates {(0,0) (1,0) (2,0) (3,0) (4,0) (5,1842.77)};
        \addplot+[fill=DodgerBlue!10, draw=black, postaction={pattern=north east lines, pattern color=DodgerBlue}] coordinates {(0,0) (1,0) (2,0) (3,0) (4,0) (5,777.42)};
        \end{axis}
        
        \begin{axis}[
            at={(plot441.east)},
            anchor=west,
            xshift=\subplotsep,
            width=0.42\textwidth,
            height=\plotmainheight*0.5,
            axis y line*=right,
            ymin=2500, ymax=5500,
            ylabel=Finetune Throughput (tokens/s),
            y tick label style={rotate=90},
            xtick={2.5},
            xmin=-1, xmax=6,
            legend=none,
            hide x axis
          ]
        
        \addplot[mark=-, mark options={scale=3}, black, thick] coordinates {(0,4639.4)};
        \addplot[mark=-, mark options={scale=3}, black, thick] coordinates {(1,4029.67)};
        \addplot[mark=-, mark options={scale=3}, black, thick] coordinates {(2,3835.22)};
        \addplot[mark=-, mark options={scale=3}, black, thick] coordinates {(4,4749.16)};
        \addplot[mark=-, mark options={scale=3}, black, thick] coordinates {(5,4445.48)};
        
        \addplot[mark=*, mark options={scale=1.25}, DarkViolet!80, thick] coordinates {(0,4639.4)};
        \addplot[mark=*, mark options={scale=1.25, fill=white, draw=DarkViolet}, DarkViolet, thick] coordinates {(1,4029.67)};
        \addplot[mark=square*, mark options={scale=1.25}, DarkOrange!80, thick] coordinates {(2,3835.22)};
        \addplot[only marks, mark=x, mark options={scale=1.5}, Red, thick] coordinates {(3,2650)};
        \addplot[mark=triangle*, mark options={scale=1.25}, DodgerBlue!80, thick] coordinates {(4,4749.16)};
        \addplot[mark=triangle*, mark options={scale=1.25, fill=white, draw=DodgerBlue}, DodgerBlue, thick] coordinates {(5,4445.48)};
        \end{axis}
        
        \begin{axis}[
            at={(plot441.below south west)},
            anchor=north west,
            yshift=-1cm,
            width=0.9\textwidth,
            height=\plotmainheight*0.3,
            ymin=0, ymax=1,
            xmin=0, xmax=1,
            axis line style={draw=none},
            xtick=\empty,
            ytick=\empty,
            legend style={
                nodes={scale=0.9, transform shape},
                legend columns=3,
                legend pos=north west,
                at={(0.5, 0.7)},
                anchor=center
            }
        ]
        
        \addlegendimage{area legend, fill=DarkViolet!30}
        \addlegendentry{}
        \addlegendimage{DarkViolet!80, mark=*, thick}
        \addlegendentry{Loquetier (Partial, Training)}
        \addlegendimage{area legend, fill=DarkViolet!10}
        \addlegendentry{Loquetier (Partial, Evaluation)}
        \addlegendimage{area legend, fill=DarkViolet!30, postaction={pattern=north east lines, pattern color=DarkViolet}}
        \addlegendentry{}
        \addlegendimage{DarkViolet, mark=*, mark options={fill=white, draw=DarkViolet}, thick}
        \addlegendentry{Loquetier (Full, Training)}
        \addlegendimage{area legend, fill=DarkViolet!10, postaction={pattern=north east lines, pattern color=DarkViolet}}
        \addlegendentry{Loquetier (Full, Evaluation)}
        \addlegendimage{area legend, fill=DarkOrange!30}
        \addlegendentry{}
        \addlegendimage{DarkOrange!80, mark=square*, thick}
        \addlegendentry{FlexLLM (Partial, Training)}
        \addlegendimage{area legend, fill=DarkOrange!10}
        \addlegendentry{FlexLLM (Partial, Evaluation)}
        \addlegendimage{area legend, fill=DarkOrange!30, postaction={pattern=north east lines, pattern color=DarkOrange}}
        \addlegendentry{}
        \addlegendimage{DarkOrange, mark=square*, mark options={fill=white, draw=DarkOrange}, thick}
        \addlegendentry{FlexLLM (Full, Training)}
        \addlegendimage{area legend, fill=DarkOrange!10, postaction={pattern=north east lines, pattern color=DarkOrange}}
        \addlegendentry{FlexLLM (Full, Evaluation)}
        \addlegendimage{area legend, fill=DodgerBlue!30}
        \addlegendentry{}
        \addlegendimage{DodgerBlue!80, mark=triangle*, thick}
        \addlegendentry{PEFT (Partial, Training)}
        \addlegendimage{area legend, fill=DodgerBlue!10}
        \addlegendentry{PEFT (Partial, Evaluation)}
        \addlegendimage{area legend, fill=DodgerBlue!30, postaction={pattern=north east lines, pattern color=DodgerBlue}}
        \addlegendentry{}
        \addlegendimage{DodgerBlue, mark=triangle*, mark options={fill=white, draw=DodgerBlue}, thick}
        \addlegendentry{PEFT (Full, Training)}
        \addlegendimage{area legend, fill=DodgerBlue!10, postaction={pattern=north east lines, pattern color=DodgerBlue}}
        \addlegendentry{PEFT (Full, Evaluation)}
        
        \end{axis}
    
    \end{tikzpicture}
    \caption{Comparison of the performance of Loquetier, FlexLLM and PEFT in fine-tuning tasks. The meanings of Partial and Full are the same as in Figure~\ref{fig:test-infer}. $\times$ indicates that the results were not obtained: FlexLLM does not support backward propagation computations for modules other than up, gate and down. PEFT can only finetune one LoRA adapter at a time, so its time cost is cumulative.}
    \label{fig:test-ft}
\end{figure}

%% file: figures/figure4.tex
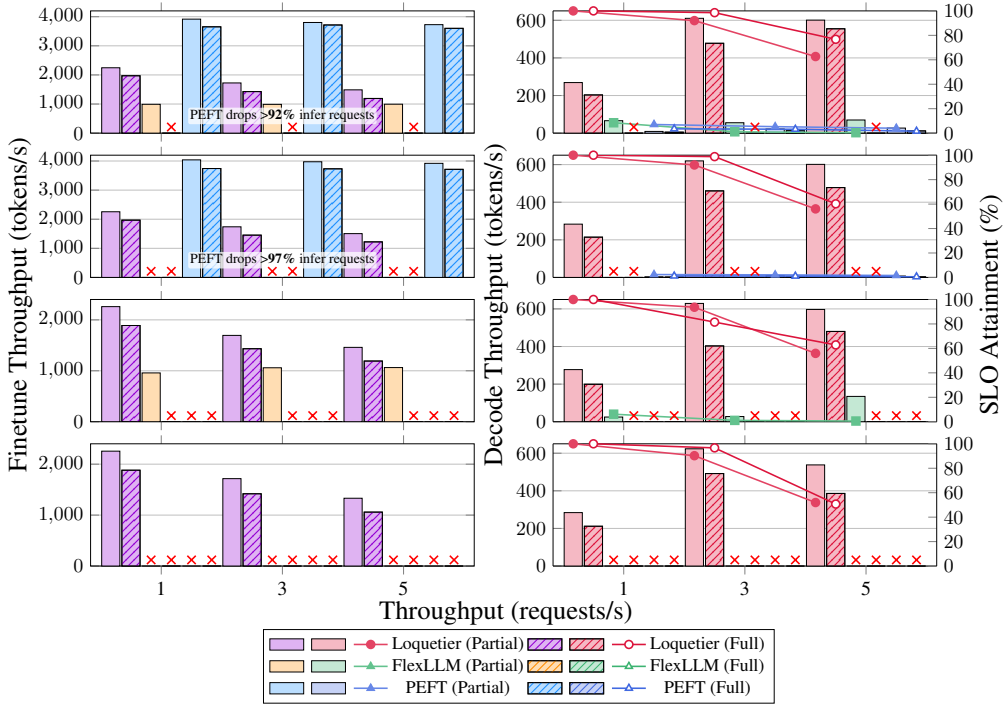
\begin{figure}
    \centering
    \begin{tikzpicture}[scale=0.75]
        \def \subplotsep {0.1\textwidth}
        \def \plotmainheight {15cm}
        \def \labeloffsetx {0.2cm}
        \def \labeloffsety {-0.2cm}
        \begin{axis}[
            name=plot451,
            width=0.6\textwidth,
            height=\plotmainheight*0.25,
            ybar,
            bar width=0.9,
            bar shift=0,
            ymajorgrids=true,
            ymin=0, ymax=4200,
            xtick={2.5,8.5,14.5},
            xticklabels=\empty,
            xmin=-1, xmax=18,
            legend=none
        ]
        
        \addplot[fill=DarkViolet!30] coordinates {(0,2246.14) (6,1724.66) (12,1487.4)};
        \addplot[fill=DarkViolet!30, postaction={pattern=north east lines, pattern color=DarkViolet}] coordinates {(1,1971.38) (7,1427.40) (13,1191.48)};
        \addplot[fill=DarkOrange!30] coordinates {(2,990.58) (8,988.42) (14,994)};
        \addplot[fill=DarkOrange!30, postaction={pattern=north east lines, pattern color=DarkViolet}] coordinates {(3,0) (9,0) (15,0)};
        \addplot[fill=DodgerBlue!30] coordinates {(4,3916.88) (10,3802.13) (16,3727.26)};
        \addplot[fill=DodgerBlue!30, postaction={pattern=north east lines, pattern color=DodgerBlue}] coordinates {(5,3653.09) (11,3716.02) (17,3600.67)};
        \node[above, fill=white, opacity=0.75, text opacity=1, font=\scriptsize, inner sep=1pt, rounded corners=1pt] at (axis description cs:0.5,0.07) {PEFT drops >\textbf{92\%} infer requests};
        \end{axis}

        \begin{axis}[
            width=0.6\textwidth,
            height=\plotmainheight*0.25,
            axis y line*=right,
            ymin=0, ymax=100,
            xtick={2.5,8.5,14.5},
            xmin=-1, xmax=18,
            legend=none,
            hide x axis,
            hide y axis
          ]
        
        \addplot[only marks, mark=x, mark options={scale=1.5}, Red, thick] coordinates {(3,5) (9,5) (15,5)};
        \end{axis}
        
        \begin{axis}[
            name=plot452,
            at={(plot451.east)},
            anchor=west,
            xshift=\subplotsep,
            width=0.6\textwidth,
            height=\plotmainheight*0.25,
            ybar,
            axis y line*=left,
            bar width=0.9,
            bar shift=0,
            ymajorgrids=true,
            ymin=0, ymax=650,
            xtick={2.5,8.5,14.5},
            xticklabels=\empty,
            xmin=-1, xmax=18,
            legend=none
        ]
        
        \addplot[fill=Crimson!30] coordinates {(0,268.90) (6,610.54) (12,600.87)};
        \addplot[fill=Crimson!30, postaction={pattern=north east lines, pattern color=Crimson}] coordinates {(1,203.08) (7,477.74) (13,554.91)};
        \addplot[fill=MediumSeaGreen!30] coordinates {(2,66.13) (8,55.32) (14,69.88)};
        \addplot[fill=MediumSeaGreen!30, postaction={pattern=north east lines, pattern color=MediumSeaGreen}] coordinates {(3,0) (9,0) (15,0)};
        \addplot[fill=RoyalBlue!30] coordinates {(4,9.09) (10,21.96) (16,26.95)};
        \addplot[fill=RoyalBlue!30, postaction={pattern=north east lines, pattern color=RoyalBlue}] coordinates {(5,5.15) (11,12.51) (17,11.81)};
        \end{axis}
        
        \begin{axis}[
            at={(plot451.east)},
            anchor=west,
            xshift=\subplotsep,
            width=0.6\textwidth,
            height=\plotmainheight*0.25,
            axis y line*=right,
            ymin=0, ymax=100,
            xtick={2.5,8.5,14.5},
            xmin=-1, xmax=18,
            legend=none,
            hide x axis
          ]
        
        \addplot[mark=*, Crimson!80, thick] coordinates {(0,100) (6,92.06) (12,62.67)};
        \addplot[mark=*, mark options={fill=white, draw=Crimson}, Crimson, thick] coordinates {(1,100) (7,98.56) (13,76.77)};
        \addplot[mark=square*, MediumSeaGreen!80, thick] coordinates {(2,8.5) (8,1.11) (14,0.37)};
        \addplot[only marks, mark=x, mark options={scale=1.5}, Red, thick] coordinates {(3,5) (9,5) (15,5)};
        \addplot[mark=triangle*, RoyalBlue!80, thick] coordinates {(4,7.06) (10,5.29) (16,4.01)};
        \addplot[mark=triangle*, mark options={fill=white, draw=RoyalBlue}, RoyalBlue, thick] coordinates {(5,3.52) (11,3.17) (17,1.6)};
        \end{axis}
        
        \begin{axis}[
            name=plot453,
            at={(plot451.below south west)},
            anchor=north west,
            width=0.6\textwidth,
            height=\plotmainheight*0.25,
            ybar,
            bar width=0.9,
            bar shift=0,
            ymajorgrids=true,
            ymin=0, ymax=4200,
            xtick={2.5,8.5,14.5},
            xticklabels=\empty,
            xmin=-1, xmax=18,
            legend=none
        ]
        
        \addplot[fill=DarkViolet!30] coordinates {(0,2256.49) (6,1739.41) (12,1505.21)};
        \addplot[fill=DarkViolet!30, postaction={pattern=north east lines, pattern color=DarkViolet}] coordinates {(1,1968.21) (7,1452.95) (13,1221.61)};
        \addplot[fill=DarkOrange!30] coordinates {(2,0) (8,0) (14,0)};
        \addplot[fill=DarkOrange!30, postaction={pattern=north east lines, pattern color=DarkViolet}] coordinates {(3,0) (9,0) (15,0)};
        \addplot[fill=DodgerBlue!30] coordinates {(4,4039.2) (10,3974.24) (16,3920.17)};
        \addplot[fill=DodgerBlue!30, postaction={pattern=north east lines, pattern color=DodgerBlue}] coordinates {(5,3741.94) (11,3731.6) (17,3713.89)};
        \node[above, fill=white, opacity=0.75, text opacity=1, font=\scriptsize, inner sep=1pt, rounded corners=1pt] at (axis description cs:0.5,0.07) {PEFT drops >\textbf{97\%} infer requests};
        \end{axis}

        \begin{axis}[
            at={(plot451.below south west)},
            anchor=north west,
            width=0.6\textwidth,
            height=\plotmainheight*0.25,
            axis y line*=right,
            ymin=0, ymax=100,
            xtick={2.5,8.5,14.5},
            xmin=-1, xmax=18,
            legend=none,
            hide x axis,
            hide y axis
          ]

        \addplot[only marks, mark=x, mark options={scale=1.5}, Red, thick] coordinates {(2,5) (8,5) (14,5)};
        \addplot[only marks, mark=x, mark options={scale=1.5}, Red, thick] coordinates {(3,5) (9,5) (15,5)};
        \end{axis}
        
        \begin{axis}[
            name=plot454,
            at={(plot453.east)},
            anchor=west,
            xshift=\subplotsep,
            width=0.6\textwidth,
            height=\plotmainheight*0.25,
            ybar,
            axis y line*=left,
            bar width=0.9,
            bar shift=0,
            ymajorgrids=true,
            ymin=0, ymax=650,
            xtick={2.5,8.5,14.5},
            xticklabels=\empty,
            xmin=-1, xmax=18,
            legend=none
        ]
        
        \addplot[fill=Crimson!30] coordinates {(0,283.37) (6,619.66) (12,601.2)};
        \addplot[fill=Crimson!30, postaction={pattern=north east lines, pattern color=Crimson}] coordinates {(1,214.13) (7,460.19) (13,477.79)};
        \addplot[fill=MediumSeaGreen!30] coordinates {(2,0) (8,0) (14,0)};
        \addplot[fill=MediumSeaGreen!30, postaction={pattern=north east lines, pattern color=MediumSeaGreen}] coordinates {(3,0) (9,0) (15,0)};
        \addplot[fill=RoyalBlue!30] coordinates {(4,3.18) (10,8.12) (16,7.52)};
        \addplot[fill=RoyalBlue!30, postaction={pattern=north east lines, pattern color=RoyalBlue}] coordinates {(5,1.69) (11,3.44) (17,2.76)};
        \end{axis}
        
        \begin{axis}[
            at={(plot453.east)},
            anchor=west,
            xshift=\subplotsep,
            width=0.6\textwidth,
            height=\plotmainheight*0.25,
            axis y line*=right,
            ymin=0, ymax=100,
            xtick={2.5,8.5,14.5},
            xmin=-1, xmax=18,
            legend=none,
            hide x axis
          ]
        
        \addplot[mark=*, Crimson!80, thick] coordinates {(0,100) (6,92) (12,56.07)};
        \addplot[mark=*, mark options={fill=white, draw=Crimson}, Crimson, thick] coordinates {(1,100) (7,98.83) (13,60.23)};
        \addplot[only marks, mark=x, mark options={scale=1.5}, Red, thick] coordinates {(2,5) (8,5) (14,5)};
        \addplot[only marks, mark=x, mark options={scale=1.5}, Red, thick] coordinates {(3,5) (9,5) (15,5)};
        \addplot[mark=triangle*, RoyalBlue!80, thick] coordinates {(4,2.34) (10,2.11) (16,1.6)};
        \addplot[mark=triangle*, mark options={fill=white, draw=RoyalBlue}, RoyalBlue, thick] coordinates {(5,1.17) (11,1.06) (17,0.53)};
        \end{axis}
        
        \begin{axis}[
            name=plot455,
            at={(plot453.below south west)},
            anchor=north west,
            width=0.6\textwidth,
            height=\plotmainheight*0.25,
            ybar,
            bar width=0.9,
            bar shift=0,
            ymajorgrids=true,
            ymin=0, ymax=2400,
            xtick={2.5,8.5,14.5},
            xticklabels=\empty,
            xmin=-1, xmax=18,
            legend=none
        ]
        
        \addplot[fill=DarkViolet!30] coordinates {(0,2259.84) (6,1695) (12,1459.95)};
        \addplot[fill=DarkViolet!30, postaction={pattern=north east lines, pattern color=DarkViolet}] coordinates {(1,1889.4) (7,1433.23) (13,1191.62)};
        \addplot[fill=DarkOrange!30] coordinates {(2,958.38) (8,1059.39) (14,1063.36)};
        \addplot[fill=DarkOrange!30, postaction={pattern=north east lines, pattern color=DarkViolet}] coordinates {(3,0) (9,0) (15,0)};
        \addplot[fill=DodgerBlue!30] coordinates {(4,0) (10,0) (16,0)};
        \addplot[fill=DodgerBlue!30, postaction={pattern=north east lines, pattern color=DodgerBlue}] coordinates {(5,0) (11,0) (17,0)};
        \end{axis}
        
        \begin{axis}[
            at={(plot453.below south west)},
            anchor=north west,
            width=0.6\textwidth,
            height=\plotmainheight*0.25,
            axis y line*=right,
            ymin=0, ymax=100,
            xtick={2.5,8.5,14.5},
            xmin=-1, xmax=18,
            legend=none,
            hide x axis,
            hide y axis
          ]
        
        \addplot[only marks, mark=x, mark options={scale=1.5}, Red, thick] coordinates {(3,5) (9,5) (15,5)};
        \addplot[only marks, mark=x, mark options={scale=1.5}, Red, thick] coordinates {(4,5) (10,5) (16,5)};
        \addplot[only marks, mark=x, mark options={scale=1.5}, Red, thick] coordinates {(5,5) (11,5) (17,5)};
        \end{axis}
        
        \begin{axis}[
            name=plot456,
            at={(plot455.east)},
            anchor=west,
            xshift=\subplotsep,
            width=0.6\textwidth,
            height=\plotmainheight*0.25,
            ybar,
            axis y line*=left,
            bar width=0.9,
            bar shift=0,
            ymajorgrids=true,
            ymin=0, ymax=650,
            xtick={2.5,8.5,14.5},
            xticklabels=\empty,
            xmin=-1, xmax=18,
            legend=none
        ]
        
        \addplot[fill=Crimson!30] coordinates {(0,277.19) (6,629.08) (12,597.02)};
        \addplot[fill=Crimson!30, postaction={pattern=north east lines, pattern color=Crimson}] coordinates {(1,199.28) (7,403.01) (13,480.15)};
        \addplot[fill=MediumSeaGreen!30] coordinates {(2,25) (8,27.63) (14,134.72)};
        \addplot[fill=MediumSeaGreen!30, postaction={pattern=north east lines, pattern color=MediumSeaGreen}] coordinates {(3,0) (9,0) (15,0)};
        \addplot[fill=RoyalBlue!30] coordinates {(4,0) (10,0) (16,0)};
        \addplot[fill=RoyalBlue!30, postaction={pattern=north east lines, pattern color=RoyalBlue}] coordinates {(5,0) (11,0) (17,0)};
        \end{axis}
        
        \begin{axis}[
            at={(plot455.east)},
            anchor=west,
            xshift=\subplotsep,
            width=0.6\textwidth,
            height=\plotmainheight*0.25,
            axis y line*=right,
            ymin=0, ymax=100,
            xtick={2.5,8.5,14.5},
            xmin=-1, xmax=18,
            legend=none,
            hide x axis
          ]
        
        \addplot[mark=*, Crimson!80, thick] coordinates {(0,100) (6,93.67) (12,55.93)};
        \addplot[mark=*, mark options={fill=white, draw=Crimson}, Crimson, thick] coordinates {(1,100) (7,81.5) (13,62.83)};
        \addplot[mark=square*, MediumSeaGreen!80, thick] coordinates {(2,6.17) (8,1.11) (14,0.57)};
        \addplot[only marks, mark=x, mark options={scale=1.5}, Red, thick] coordinates {(3,5) (9,5) (15,5)};
        \addplot[only marks, mark=x, mark options={scale=1.5}, Red, thick] coordinates {(4,5) (10,5) (16,5)};
        \addplot[only marks, mark=x, mark options={scale=1.5}, Red, thick] coordinates {(5,5) (11,5) (17,5)};
        \end{axis}
        
        \begin{axis}[
            name=plot457,
            at={(plot455.below south west)},
            anchor=north west,
            width=0.6\textwidth,
            height=\plotmainheight*0.25,
            ybar,
            bar width=0.9,
            bar shift=0,
            ymajorgrids=true,
            ymin=0, ymax=2400,
            xtick={2.5,8.5,14.5},
            xticklabels={1,3,5},
            xmin=-1, xmax=18,
            legend=none
        ]
        
        \addplot[fill=DarkViolet!30] coordinates {(0,2255.36) (6,1715.88) (12,1331.1)};
        \addplot[fill=DarkViolet!30, postaction={pattern=north east lines, pattern color=DarkViolet}] coordinates {(1,1881.48) (7,1418.7) (13,1058.8)};
        \addplot[fill=DarkOrange!30] coordinates {(2,0) (8,0) (14,0)};
        \addplot[fill=DarkOrange!30, postaction={pattern=north east lines, pattern color=DarkViolet}] coordinates {(3,0) (9,0) (15,0)};
        \addplot[fill=DodgerBlue!30] coordinates {(4,0) (10,0) (16,0)};
        \addplot[fill=DodgerBlue!30, postaction={pattern=north east lines, pattern color=DodgerBlue}] coordinates {(5,0) (11,0) (17,0)};
        \end{axis}
        
        \begin{axis}[
            at={(plot455.below south west)},
            anchor=north west,
            width=0.6\textwidth,
            height=\plotmainheight*0.25,
            axis y line*=right,
            ymin=0, ymax=100,
            xtick={2.5,8.5,14.5},
            xmin=-1, xmax=18,
            legend=none,
            hide x axis,
            hide y axis
          ]
        
        \addplot[only marks, mark=x, mark options={scale=1.5}, Red, thick] coordinates {(2,5) (8,5) (14,5)};
        \addplot[only marks, mark=x, mark options={scale=1.5}, Red, thick] coordinates {(3,5) (9,5) (15,5)};
        \addplot[only marks, mark=x, mark options={scale=1.5}, Red, thick] coordinates {(4,5) (10,5) (16,5)};
        \addplot[only marks, mark=x, mark options={scale=1.5}, Red, thick] coordinates {(5,5) (11,5) (17,5)};
        \end{axis}
        
        \begin{axis}[
            name=plot458,
            at={(plot457.east)},
            anchor=west,
            xshift=\subplotsep,
            width=0.6\textwidth,
            height=\plotmainheight*0.25,
            ybar,
            axis y line*=left,
            bar width=0.9,
            bar shift=0,
            ymajorgrids=true,
            ymin=0, ymax=650,
            xtick={2.5,8.5,14.5},
            xticklabels={1,3,5},
            xmin=-1, xmax=18,
            legend=none
        ]
        
        \addplot[fill=Crimson!30] coordinates {(0,284.15) (6,623.51) (12,537.8)};
        \addplot[fill=Crimson!30, postaction={pattern=north east lines, pattern color=Crimson}] coordinates {(1,211.61) (7,491.79) (13,386.1)};
        \addplot[fill=MediumSeaGreen!30] coordinates {(2,0) (8,0) (14,0)};
        \addplot[fill=MediumSeaGreen!30, postaction={pattern=north east lines, pattern color=MediumSeaGreen}] coordinates {(3,0) (9,0) (15,0)};
        \addplot[fill=RoyalBlue!30] coordinates {(4,0) (10,0) (16,0)};
        \addplot[fill=RoyalBlue!30, postaction={pattern=north east lines, pattern color=RoyalBlue}] coordinates {(5,0) (11,0) (17,0)};
        \end{axis}
        
        \begin{axis}[
            at={(plot457.east)},
            anchor=west,
            xshift=\subplotsep,
            width=0.6\textwidth,
            height=\plotmainheight*0.25,
            axis y line*=right,
            ymin=0, ymax=100,
            xtick={2.5,8.5,14.5},
            xmin=-1, xmax=18,
            legend=none,
            hide x axis
          ]
        
        \addplot[mark=*, Crimson!80, thick] coordinates {(0,100) (6,90.28) (12,51.97)};
        \addplot[mark=*, mark options={fill=white, draw=Crimson}, Crimson, thick] coordinates {(1,100) (7,96.67) (13,50.7)};
        \addplot[only marks, mark=x, mark options={scale=1.5}, Red, thick] coordinates {(2,5) (8,5) (14,5)};
        \addplot[only marks, mark=x, mark options={scale=1.5}, Red, thick] coordinates {(3,5) (9,5) (15,5)};
        \addplot[only marks, mark=x, mark options={scale=1.5}, Red, thick] coordinates {(4,5) (10,5) (16,5)};
        \addplot[only marks, mark=x, mark options={scale=1.5}, Red, thick] coordinates {(5,5) (11,5) (17,5)};
        \end{axis}
        
        \path (current bounding box.north) -- (current bounding box.south)
            coordinate[midway] (vertical_center);
            
        \node[anchor=center] at ($(current bounding box.south) + (0,\labeloffsety)$)
            {Throughput (requests/s)};
        
        \node[rotate=90, anchor=center] at ($(current bounding box.west) - (\labeloffsetx,\labeloffsety)$)
            {Finetune Throughput (tokens/s)};
        
        \node[rotate=90, anchor=center] at ($(current bounding box.east) - (\labeloffsetx+0.6\textwidth,\labeloffsety)$)
            {Decode Throughput (tokens/s)};
        
        \node[rotate=90, anchor=center] at ($(current bounding box.east) - (-\labeloffsetx,\labeloffsety)$)
            {SLO Attainment (\%)};
        
        \begin{axis}[
            at={(plot457.below south)},
            anchor=north,
            xshift=0.3\textwidth,
            width=0.9\textwidth,
            height=\plotmainheight*0.12,
            ymin=0, ymax=1,
            xmin=0, xmax=1,
            axis line style={draw=none},
            xtick=\empty,
            ytick=\empty,
            legend style={
                nodes={scale=0.9, transform shape},
                legend columns=6,
                legend pos=north west,
                at={(0.5, -4.25)},
                anchor=center
            }
        ]
        
        \addlegendimage{area legend, fill=DarkViolet!30}
        \addlegendentry{}
        \addlegendimage{area legend, fill=Crimson!30}
        \addlegendentry{}
        \addlegendimage{Crimson!80, mark=*, thick}
        \addlegendentry{Loquetier (Partial)}
        \addlegendimage{area legend, fill=DarkViolet!30, postaction={pattern=north east lines, pattern color=DarkViolet}}
        \addlegendentry{}
        \addlegendimage{area legend, fill=Crimson!30, postaction={pattern=north east lines, pattern color=Crimson}}
        \addlegendentry{}
        \addlegendimage{Crimson, mark=*, mark options={fill=white, draw=Crimson}, thick}
        \addlegendentry{Loquetier (Full)}
        \addlegendimage{area legend, fill=DarkOrange!30}
        \addlegendentry{}
        \addlegendimage{area legend, fill=MediumSeaGreen!30}
        \addlegendentry{}
        \addlegendimage{MediumSeaGreen!80, mark=triangle*, thick}
        \addlegendentry{FlexLLM (Partial)}
        \addlegendimage{area legend, fill=DarkOrange!30, postaction={pattern=north east lines, pattern color=DarkOrange}}
        \addlegendentry{}
        \addlegendimage{area legend, fill=MediumSeaGreen!30, postaction={pattern=north east lines, pattern color=MediumSeaGreen}}
        \addlegendentry{}
        \addlegendimage{MediumSeaGreen, mark=triangle*, mark options={fill=white, draw=MediumSeaGreen}, thick}
        \addlegendentry{FlexLLM (Full)}
        \addlegendimage{area legend, fill=DodgerBlue!30}
        \addlegendentry{}
        \addlegendimage{area legend, fill=RoyalBlue!30}
        \addlegendentry{}
        \addlegendimage{RoyalBlue!80, mark=triangle*, thick}
        \addlegendentry{PEFT (Partial)}
        \addlegendimage{area legend, fill=DodgerBlue!30, postaction={pattern=north east lines, pattern color=DodgerBlue}}
        \addlegendentry{}
        \addlegendimage{area legend, fill=RoyalBlue!30, postaction={pattern=north east lines, pattern color=RoyalBlue}}
        \addlegendentry{}
        \addlegendimage{RoyalBlue, mark=triangle*, mark options={fill=white, draw=RoyalBlue}, thick}
        \addlegendentry{PEFT (Full)}
        
        \end{axis}
    
    \end{tikzpicture}
    \caption{Comparison of the performance of Loquetier and PEFT in unified tasks. The 4 subplots correspond respectively to single-finetune \& single-infer, single-finetune \& multi-infer, multi-finetune \& single-infer, and multi-finetune \& multi-infer. The meanings of Partial and Full are the same as in Figure~\ref{fig:test-infer}. $\times$ indicates that the results were not obtained: FlexLLM and PEFT can only finetune 1 LoRA at a time due to GPU memory limitations, causing it to fail the multi-LoRA fine-tuning scenarios; FlexLLM only support 3 target modules as mentioned in previous figures.
}
    \label{fig:test-unified}
\end{figure}

%% file: figures/figure5.tex
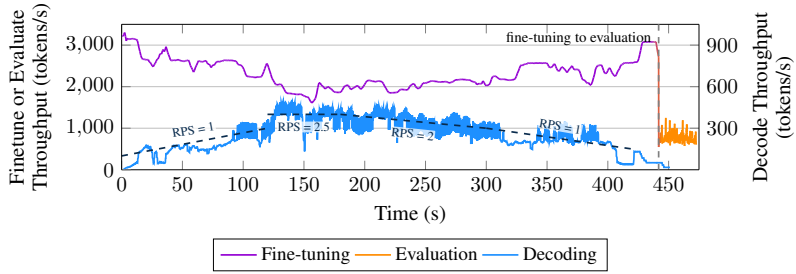
\begin{figure}
    \centering
    \begin{tikzpicture}[scale=0.8]
        \def \plotmainheight {4cm}
        
        \begin{axis}[
            name=plot461,
            width=0.8\textwidth,
            height=\plotmainheight,
            axis y line*=right,
            bar width=0.9,
            bar shift=0,
            ymajorgrids=true,
            ymin=0, ymax=1050,
            ylabel style={align=center},
            ylabel=Decode Throughput\\(tokens/s),
            ytick={300,600,900},
            xmin=0, xmax=475,
            xlabel=Time (s),
            legend=none
        ]
        
        \addplot+[
            mark=none, DodgerBlue, thick
        ] table [x index=0, y index=1, col sep=space] {results/Mutable_Inference.txt};

        \draw[dashed, DodgerBlue!30!Black, thick] (axis cs:0,100) -- (axis cs:120,300)
            node[pos=0.5, above, yshift=3pt, rotate=10, fill=white, opacity=0.75, text opacity=1, inner sep=0.5pt, rounded corners=0.5pt, font=\tiny] {RPS = 1};
        \draw[dashed, DodgerBlue!30!Black, thick] (axis cs:120,400) -- (axis cs:180,400)
            node[pos=0.5, below, yshift=-3pt, rotate=0, fill=white, opacity=0.75, text opacity=1, inner sep=0.5pt, rounded corners=0.5pt, font=\tiny] {RPS = 2.5};
        \draw[dashed, DodgerBlue!30!Black, thick] (axis cs:180,400) -- (axis cs:300,300)
            node[pos=0.5, below, yshift=-3pt, rotate=-6, fill=white, opacity=0.75, text opacity=1, inner sep=0.5pt, rounded corners=0.5pt, font=\tiny] {RPS = 2};
        \draw[dashed, DodgerBlue!30!Black, thick] (axis cs:300,300) -- (axis cs:420,150)
            node[pos=0.5, above, yshift=3pt, rotate=-10, fill=white, opacity=0.75, text opacity=1, inner sep=0.5pt, rounded corners=0.5pt, font=\tiny] {RPS = 1};
        
        \end{axis}

        \begin{axis}[
            width=0.8\textwidth,
            height=\plotmainheight,
            ybar,
            axis y line*=left,
            bar width=0.9,
            bar shift=0,
            ymin=0, ymax=3500,
            ylabel style={align=center},
            ylabel=Finetune or Evaluate\\Throughput (tokens/s),
            xmin=0, xmax=475,
            legend=none,
            colormap={custommap}{
                color(0cm)=(DarkViolet);
                color(1cm)=(DarkOrange)
            },
            hide x axis
        ]
        
        \addplot+[
            thick, mesh, point meta={
                x > 445 ? 1 : (x > 435 ? (x-435)/10 : 0)
            }, colormap name=custommap, shader=interp
        ] table [x index=0, y index=1, col sep=space] {results/Mutable_Finetuning.txt};
        
        \draw[dashed, gray, thick] (axis cs:441.8,3500) -- (axis cs:441.8,0);
        
        \node[below left, font=\scriptsize] at (axis cs:441.8,3500) {fine-tuning to evaluation};
        
        \end{axis}
        
        \begin{axis}[
            at={(plot461.below south west)},
            anchor=north west,
            yshift=-0.4cm,
            width=0.8\textwidth,
            height=\plotmainheight*0.4,
            ymin=0, ymax=1,
            xmin=0, xmax=1,
            axis line style={draw=none},
            xtick=\empty,
            ytick=\empty,
            xlabel style={font=\tiny},
            ylabel style={font=\tiny},
            legend style={
                nodes={scale=0.9, transform shape},
                legend columns=4,
                legend pos=north west,
                at={(0.5, 0.7)},
                anchor=center
            }
        ]
        
        \addlegendimage{DarkViolet, mark=none, thick}
        \addlegendentry{Fine-tuning}
        \addlegendimage{DarkOrange, mark=none, thick}
        \addlegendentry{Evaluation}
        \addlegendimage{DodgerBlue, mark=none, thick}
        \addlegendentry{Decoding}
        
        \end{axis}
    
    \end{tikzpicture}
    \caption{Performance of Loquetier under dynamic load in unified task.}
    \label{fig:test-mutable}
\end{figure}

%% file: figures/figure6.tex
\begin{figure}
    \centering
    \begin{tikzpicture}[scale=0.8]
        \def \plotmainheight {4cm}
        
        \begin{axis}[
            name=plot461,
            width=0.8\textwidth,
            height=\plotmainheight,
            axis y line*=right,
            bar width=0.9,
            bar shift=0,
            ymajorgrids=true,
            ymin=0, ymax=1050,
            ylabel style={align=center},
            ylabel=Decode Throughput\\(tokens/s),
            ytick={300,600,900},
            xmin=0, xmax=7100,
            xlabel=Time (s),
            legend=none
        ]
        
        \addplot+[
            mark=none, DodgerBlue, thick
        ] table [x index=0, y index=1, col sep=space] {results/BurstGPT_Workload_Inference.txt};
        
        \end{axis}

        \begin{axis}[
            width=0.8\textwidth,
            height=\plotmainheight,
            ybar,
            axis y line*=left,
            bar width=0.9,
            bar shift=0,
            ymin=0, ymax=3200,
            ylabel style={align=center},
            ylabel=Finetune or Evaluate\\Throughput (tokens/s),
            xmin=0, xmax=7300,
            legend=none,
            colormap={custommap}{
                color(0cm)=(DarkOrange);
                color(1cm)=(DarkViolet)
            },
            hide x axis
        ]
        
        \addplot+[
            thick, mesh, point meta={
                y > 1600 ? 1 : (y > 1200 ? (y-1200)/400 : 0)
            }, colormap name=custommap, shader=interp
        ] table [x index=0, y index=1, col sep=space] {results/BurstGPT_Workload_Finetuning.txt};
        
        \end{axis}
        
        \begin{axis}[
            at={(plot461.below south west)},
            anchor=north west,
            yshift=-0.4cm,
            width=0.8\textwidth,
            height=\plotmainheight*0.4,
            ymin=0, ymax=1,
            xmin=0, xmax=1,
            axis line style={draw=none},
            xtick=\empty,
            ytick=\empty,
            xlabel style={font=\tiny},
            ylabel style={font=\tiny},
            legend style={
                nodes={scale=0.9, transform shape},
                legend columns=4,
                legend pos=north west,
                at={(0.5, 0.7)},
                anchor=center
            }
        ]
        
        \addlegendimage{DarkViolet, mark=none, thick}
        \addlegendentry{Fine-tuning}
        \addlegendimage{DarkOrange, mark=none, thick}
        \addlegendentry{Evaluation}
        \addlegendimage{DodgerBlue, mark=none, thick}
        \addlegendentry{Decoding}
        
        \end{axis}
    
    \end{tikzpicture}
    \caption{Performance of Loquetier under simulated real-world load in unified task.}
    \label{fig:test-burstgpt}
\end{figure}
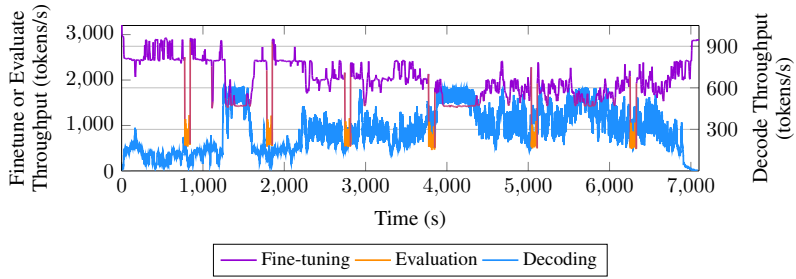

%% file: chapters/chapter5.tex
\section{Conclusion}

We present Loquetier, a virtualized multi-LoRA framework that runs fine-tuning and inference tasks uniformly. Loquetier performs well in the inference task, with an SLO attainment 20.8$\times$ higher than PEFT, and up to 3.0$\times$ that of FlexLLM at high request arrival rates, maintaining comparable efficiency in the fine-tuning task. In the unified task, inference efficiency is maintained as much as possible with an SLO attainment 46.4$\times$ higher than PEFT, well balancing the performance of fine-tuning and inference tasks.

%% file: chapters/appendix.tex
\newpage
\appendix

\section{Limitations}
\label{sec:limitations}

Loquetier can be improved by combination with other training and fine-tuning optimization methods, and by applying other cluster management systems to provide clusterized services. We plan to provide a backward propagation kernel operating in concert with the SMLM kernel to accelerate fine-tuning.

Co-operating with forward propagation computations during backward propagation can go some way to balancing the bottleneck of GPU memory bandwidth and computational resources for fine-tuning and inference tasks, and Loquetier can be further improved in this regard. We also consider providing support for other LoRA-like PEFT methods.

\section{Solutions of FlexLLM Backward Procedure Issues}
\label{sec:solutions}

Several cases of operation remain unimplemented in FlexLLM's gradient computation logic, including OP\_GELU, OP\_RELU, OP\_SIGMOID, OP\_TANH, and OP\_ELU. We noticed that their repository contained forward and backward kernels related to these operations, but they had never applied these backward kernels to their computation flow, resulting in its inability to perform fine-tuning tasks. We ultimately built a runnable version by instantiating these kernels at the missing locations. This fix was based on our understanding of their framework, and we did not implement any additional computational steps. The correction will not result in a performance degradation of FlexLLM.

\section{Experimental Metrics}
\label{sec:metrics}

We use the following experimental metircs:

\begin{itemize}
    \item \textbf{Service Level Objective (SLO)}: Measuring service satisfaction at a level.
    \item \textbf{SLO Attainment}: Percentage of all requests reaching the given SLO.
    \item \textbf{Request Throughput}: Throughput of incoming inference requests. Measured as request per second (RPS).
    \item \textbf{Decode Throughput}: Throughput of inference requests in decoding. Measured as decode token per second (DTPS).
    \item \textbf{Finetune Throughput}: Throughput of fine-tuning requests in training forward. Measured as finetune token per second (FTPS).
    \item \textbf{Evaluate Throughput}: Throughput of fine-tuning requests in evaluation forward. Measured as evaluate token per second (ETPS).
\end{itemize}

\section{Experimental Settings}
\label{sec:settings}

\subsection{SLO}

For all inference requests in the experiments, we use the following targets in Figure~\ref{tab:metrics-slo} as SLO.

Since PEFT performs batch generation with padding, resulting in additional computational delays in prefilling and decoding, we do not require more decoding aspects for PEFT.

\begin{table}[hbp]
    \caption{SLO settings.}
    \label{tab:metrics-slo}
    \centering
    \begin{tabular}{llll}
        \toprule
        \makecell{\\Framework or System} &
        \makecell{\\Max Waiting Time (s)} &
        \makecell{Mean Decoding\\Latency (ms)} &
        \makecell{Max Decoding\\Latency (ms)} \\
        \midrule
        Loquetier &
        6 & 200 & 1,000 \\
        PEFT &
        6 & - & -       \\
        FlexLLM &
        6 & 200 & 1,000 \\
        \bottomrule
    \end{tabular}
\end{table}

\begin{table}[htbp]
    \caption{Inference-only tasks configurations. The number of requests in multiple (4) LoRAs is expressed as the total number / number of one LoRA.}
    \label{tab:metrics-infer}
    \centering
    \begin{tabular}{lllll}
        \toprule
        & \multicolumn{2}{c}{Single (1) LoRA} & \multicolumn{2}{c}{Multiple (4) LoRAs}                                                     \\
        \cmidrule(r){2-3} \cmidrule(r){4-5}
        Throughput (RPS) &
        Requests & Max New Tokens &
        Requests & Max New Tokens \\
        \midrule
        1 &
        800 & 400 & 800 / 200 & 400     \\
        2 &
        1,600 & 400 & 1,600 / 400 & 400 \\
        3 &
        2,400 & 400 & 2,400 / 600 & 400 \\
        4 &
        3,200 & 300 & 3,200 / 800 & 300 \\
        5 &
        4,000 & 200 & 4,000 / 500 & 200 \\
        \bottomrule
    \end{tabular}
\end{table}

\subsection{Inference}

We use the following configurations in Figure~\ref{tab:metrics-infer} to test the inference-only tasks.

Note that the maximum tokens supported by FlexLLM is 1024, so all inference requests for FlexLLM do not exceed this limit.

\subsection{Fine-tuning}

We use the following configurations in Figure~\ref{tab:metrics-ft} to test the fine-tuning-only tasks.

\begin{table}[htbp]
    \caption{Fine-tuning-only tasks configurations.}
    \label{tab:metrics-ft}
    \centering
    \begin{tabular}{lll}
        \toprule
        Configurations & Single (1) LoRA & Multiple (2) LoRAs \\
        \midrule
        LoRA Config & & \\
        \hspace{2em}r (rank) & 8 & 8                          \\
        \hspace{2em}lora\_alpha & 16 & 16                     \\
        \hspace{2em}lora\_dropout & 0.05 & 0.05               \\
        \hspace{2em}bias & none & none                        \\
        \hspace{2em}task\_type & CAUSAL\_LM & CAUSAL\_LM      \\
        \hspace{2em}init\_lora\_weights & gaussian & gaussian \\
        Training Args & & \\
        \hspace{2em}per\_device\_train\_batch\_size & 2 & 1   \\
        \hspace{2em}per\_device\_eval\_batch\_size & 2 & 1    \\
        \hspace{2em}num\_train\_epochs & 4 & 4                \\
        \hspace{2em}eval\_strategy & epoch & epoch            \\
        \hspace{2em}logging\_strategy & steps & steps         \\
        \hspace{2em}logging\_steps & 100 & 100                \\
        \hspace{2em}save\_strategy & epoch & epoch            \\
        \hspace{2em}learning\_rate & 2e-5 & 2e-5              \\
        \hspace{2em}gradient\_accumulation\_steps & 4 & 4     \\
        \hspace{2em}report\_to & none & none                  \\
        \bottomrule
    \end{tabular}
\end{table}

\subsection{Unified fine-tuning and inference}

We use the following configurations in Figure~\ref{tab:metrics-unified} to test the unified tasks.

\begin{table}[htbp]
    \caption{Unified tasks configurations.}
    \label{tab:metrics-unified}
    \centering
    \begin{tabular}{lllll}
        \toprule
        & \multicolumn{2}{c}{Single (1) LoRA} & \multicolumn{2}{c}{Multiple (4) LoRAs} \\
        \cmidrule(r){2-3} \cmidrule(r){4-5}
        Throughput (RPS) &
        Requests & Max New Tokens &
        Requests & Max New Tokens \\
        \midrule
        1 &
        600 & 400 & 600 / 150 & 400     \\
        2 &
        1,200 & 400 & 1,200 / 300 & 400 \\
        3 &
        1,800 & 400 & 1,800 / 450 & 400 \\
        4 &
        2,400 & 300 & 2,400 / 600 & 300 \\
        5 &
        3,000 & 200 & 3,000 / 750 & 200 \\
        \bottomrule
    \end{tabular}
    \begin{tabular}{lll}
        \toprule
        Configurations & Single (1) LoRA & Multiple (2) LoRAs \\
        \midrule
        LoRA Config & & \\
        \hspace{2em}r (rank) & 8 & 8                          \\
        \hspace{2em}lora\_alpha & 16 & 16                     \\
        \hspace{2em}lora\_dropout & 0.05 & 0.05               \\
        \hspace{2em}bias & none & none                        \\
        \hspace{2em}task\_type & CAUSAL\_LM & CAUSAL\_LM      \\
        \hspace{2em}init\_lora\_weights & gaussian & gaussian \\
        Training Args & & \\
        \hspace{2em}per\_device\_train\_batch\_size & 2 & 1   \\
        \hspace{2em}per\_device\_eval\_batch\_size & 2 & 1    \\
        \hspace{2em}num\_train\_epochs & 1 & 1                \\
        \hspace{2em}eval\_strategy & epoch & epoch            \\
        \hspace{2em}logging\_strategy & steps & steps         \\
        \hspace{2em}logging\_steps & 100 & 100                \\
        \hspace{2em}save\_strategy & epoch & epoch            \\
        \hspace{2em}learning\_rate & 2e-5 & 2e-5              \\
        \hspace{2em}gradient\_accumulation\_steps & 4 & 4     \\
        \hspace{2em}report\_to & none & none                  \\
        \bottomrule
    \end{tabular}
\end{table}

Note that the maximum tokens supported by FlexLLM is 1024, so all inference requests for FlexLLM do not exceed this limit.

\subsection{Mutable capacity allocation simulation}

We use the following configurations in Figure~\ref{tab:metrics-mutable} for inference requests and the single LoRA fine-tuning configurations in Figure~\ref{tab:metrics-unified} for fine-tuning requests to test the mutable unified tasks.

\begin{table}
    \caption{Mutable unified tasks configurations.}
    \label{tab:metrics-mutable}
    \centering
    \begin{tabular}{llllll}
        \toprule
        & \multicolumn{5}{c}{Multiple (4) LoRAs} \\
        \cmidrule(r){2-6}
        Index &
        LoRA Index & Requests & Throughput (RPS) &
        Start at (s) & Duration (s) \\
        \midrule
        1 &
        0 & 120 & 1 & 0 & 120    \\
        2 &
        1 & 150 & 2.5 & 120 & 60 \\
        3 &
        2 & 240 & 2 & 180 & 120  \\
        4 &
        3 & 120 & 1 & 300 & 120  \\
        \bottomrule
    \end{tabular}
\end{table}

\subsection{Simulated real-world workload}
\label{sec:simr}

We use the following time periods in Figure~\ref{tab:metrics-burstgpt} to test (the inference task of) the simulated real-world workload. The fine-tuning task use the same configuration as mutable capacity allocation simulation.

Based on our analysis of the BurstGPT dataset, less than one-third of the time periods correspond to high-load scenarios, with the majority being low-load periods. Given the lower challenge of low-load periods, we adopt the configuration above to ensure representative coverage. Among the selected workloads, high-load periods include several minutes where the RPS exceeded 5, with a peak RPS of 11.

\begin{table}
    \caption{Time periods configurations. Peak RPS refers to the highest RPS within a 2-second interval.}
    \label{tab:metrics-burstgpt}
    \centering
    \begin{tabular}{llll}
        \toprule
        Time Period & Requests & Mean RPS & Peak RPS \\
        \midrule
        Day 29, 13:00 \textasciitilde 13:20 & 676   & 0.563 & 1.5  \\
        Day 29, 15:00 \textasciitilde 15:20 & 2,145 & 1.788 & 11.5 \\
        Day 29, 16:00 \textasciitilde 16:20 & 1,465 & 1.226 & 7    \\
        Day 33, 13:40 \textasciitilde 14:00 & 2,823 & 2.354 & 10   \\
        Day 33, 11:40 \textasciitilde 12:00 & 2,360 & 1.966 & 12   \\
        Day 33, 11:00 \textasciitilde 11:20 & 1,856 & 1.547 & 10.5 \\
        \bottomrule
    \end{tabular}
\end{table}

\section{Detailed Information Related to S-LoRA in the Experiment}
\label{sec:slora}

S-LoRA does not support the LLaMA 3 series models, and its repository has been archived. This limitation arises from the Group Query Attention (GQA) architecture used in LLaMA 3, where the shapes of K and V differ from those of Q and O. Consequently, the shape of the weight matrix B in the LoRA linear layers for K and V also differs from Q and O. Current S-LoRA requires all LoRA weights within the same layer to be concatenated at runtime. However, due to the shape discrepancies mentioned above, this concatenation operation fails. As a workaround, we replicate K and V weights in advance during model initialization to enable S-LoRA to start properly.

"Partial" in Figure~\ref{fig:test-infer} means that only 4 modules are enabled for S-LoRA including q, k, v, and o, as S-LoRA supports applying LoRA only on these 4 linear layers, and does not support the up, gate, and down layers within the MLP. Therefore, its runtime efficiency resembles the Partial scenario described in our paper, where only three linear layers (up, gate, down) are targeted.

In experiments, we observed instability in the S-LoRA kernel, which frequently produced incorrect outputs leading to NaN or Inf values. These errors propagate quickly, causing model generation failures. At this time, we did not modify the kernel. Our preliminary analysis suggests this may be due to missing synchronization mechanisms in some computational steps. (Note that this is an initial observation and may not be definitive.)

Due to these issues, S-LoRA struggled to complete all inference requests in our scenarios, as it frequently outputs the eos token directly, making SLO appear better than it actually should be.